\pdfoutput=1
\documentclass[10pt,twocolumn,letterpaper]{article}

\usepackage[pagenumbers]{cvpr} 

\usepackage{graphicx}
\usepackage{amsmath}
\usepackage{amssymb}
\usepackage{booktabs}
\usepackage[accsupp]{axessibility}  

\usepackage{makecell}
\usepackage{pifont}
\usepackage{multirow}
\usepackage{caption}
\usepackage{algorithmic,algorithm}
\usepackage[pagebackref,breaklinks,colorlinks]{hyperref}

\usepackage[capitalize]{cleveref}
\crefname{section}{Sec.}{Secs.}
\Crefname{section}{Section}{Sections}
\Crefname{table}{Table}{Tables}
\crefname{table}{Tab.}{Tabs.}

\def\ourmodel{ZBS}
\def\etc{\textit{etc}}
\def\etal{\textit{et al.}}

\def\cvprPaperID{10557} 
\def\confName{CVPR}
\def\confYear{2023}

\begin{document}

\title{{\ourmodel}: Zero-shot Background Subtraction via Instance-level Background Modeling and Foreground Selection}

\author{Yongqi An $^{1,2}$ \quad  Xu Zhao $^{1,}$\thanks{Corresponding Author}  \quad Tao Yu $^{1,2}$ \quad Haiyun Guo$^{1,2}$ \\
\quad Chaoyang Zhao $^{1}$ \quad Ming Tang $^{1,2}$\quad Jinqiao Wang $^{1,2}$\\
  National Laboratory of Pattern Recognition, Institute of Automation, CAS, Beijing, China$^{1}$ \\ 
  School of Artificial Intelligence, University of Chinese Academy of Sciences, Beijing, China$^{2}$\\
   \\
{\tt\small \{yongqi.an,xu.zhao,haiyun.guo,tangm,jqwang\}@nlpr.ia.ac.cn}\\
{\tt\small  yutao2022@ia.ac.cn}
}
\maketitle

\begin{abstract}
    Background subtraction (BGS) aims to extract all moving objects in the video frames to obtain binary foreground segmentation masks. Deep learning has been widely used in this field. Compared with supervised-based BGS methods, unsupervised methods have better generalization. However, previous unsupervised deep learning BGS algorithms perform poorly in sophisticated scenarios such as shadows or night lights, and they cannot detect objects outside the pre-defined categories. In this work, we propose an unsupervised BGS algorithm based on zero-shot object detection called Zero-shot Background Subtraction ({\ourmodel}). The proposed method fully utilizes the advantages of zero-shot object detection to build the open-vocabulary instance-level background model. Based on it, the foreground can be effectively extracted by comparing the detection results of new frames with the background model. ZBS performs well for sophisticated scenarios, and it has rich and extensible categories. Furthermore, our method can easily generalize to other tasks, such as abandoned object detection in unseen environments. We experimentally show that {\ourmodel} surpasses state-of-the-art unsupervised BGS methods by 4.70$\%$ F-Measure on the CDnet 2014 dataset. The code is released at \textit{\url{https://github.com/CASIA-IVA-Lab/ZBS}}.
\end{abstract}

\section{Introduction}
\label{sec:intro}

Background subtraction (BGS) is a fundamental task in computer vision applications~\cite{traditional_review_2014}, such as autonomous navigation, visual surveillance, human activity recognition, \etc~\cite{bgs_applications_2020}. BGS aims to extract all moving objects as foreground in each video frame and outputs binary segmentations.

\begin{figure}[t]
\centering
\includegraphics[width=1\columnwidth]{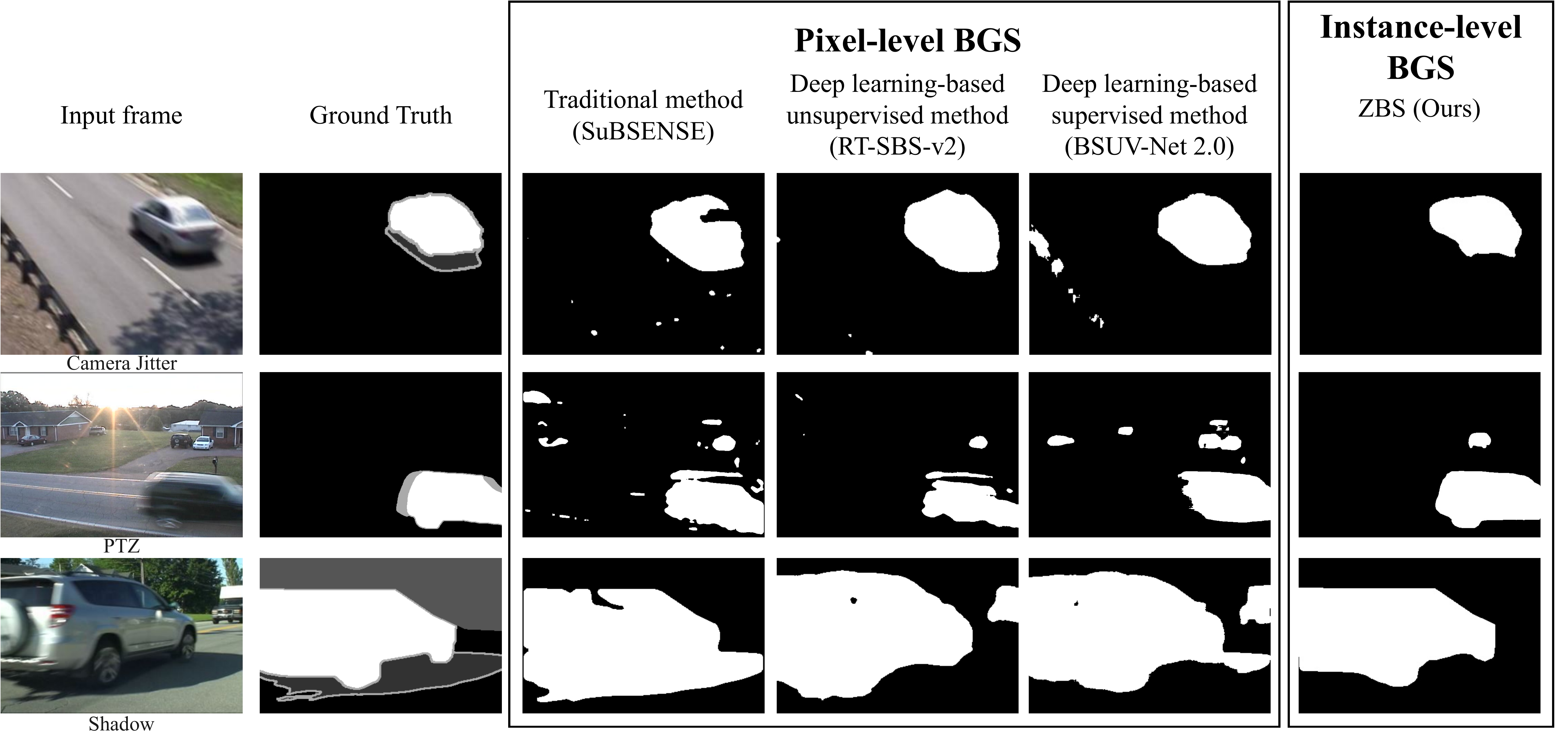} 
\caption{The performance of different BGS methods. Previous BGS methods based on pixel-level background models may misjudge noisy background as foreground objects, such as cameraJitter, PTZ, and shadow. Our method based on an instance-level background model can obtain precise foreground edges, effectively reducing the confusion of background pixels as foreground objects.}
\label{fig:pix_ins}
\end{figure}

The most straightforward BGS algorithm is to directly compare the current frame with the "stationary" background image~\cite{traditional_review_2014}. However, this strategy cannot handle complex scenarios, such as dynamic background, illumination changes, and shadows. Therefore, more sophisticated BGS techniques~\cite{traditional_review_2014, SJabri2000DetectionAL, DongxiangZhou2005ModifiedGB, FgSegNet} have been proposed in the past decades. The traditional methods improve performance in two aspects. The first is to design more robust feature representations, including color features~\cite{colorandtexturefeatures}, edge features~\cite{SJabri2000DetectionAL}, motion features~\cite{DongxiangZhou2005ModifiedGB}, and texture features~\cite{fuzzytexturefeature}. The second is to design more suitable background models, such as Gaussian mixture models~\cite{Guassion}, kernel density estimation models~\cite{kernel}, CodeBook~\cite{codebook}, ViBe~\cite{vibe}, SuBSENSE~\cite{SuBSENSE}, and PAWCS~\cite{PAWCS}. The traditional methods have relatively adequate generalization capacity since they are not optimized on specific scenarios or categories of objects. However, these methods only utilize hand-craft features to determine whether each pixel belongs to the foreground. We call these methods pixel-level BGS since they use pixel-based or local pixels-based background models. They are sensitive to natural variations such as lighting and weather. 

\begin{figure*}[htb]
\centering
\includegraphics[width=0.83\linewidth]{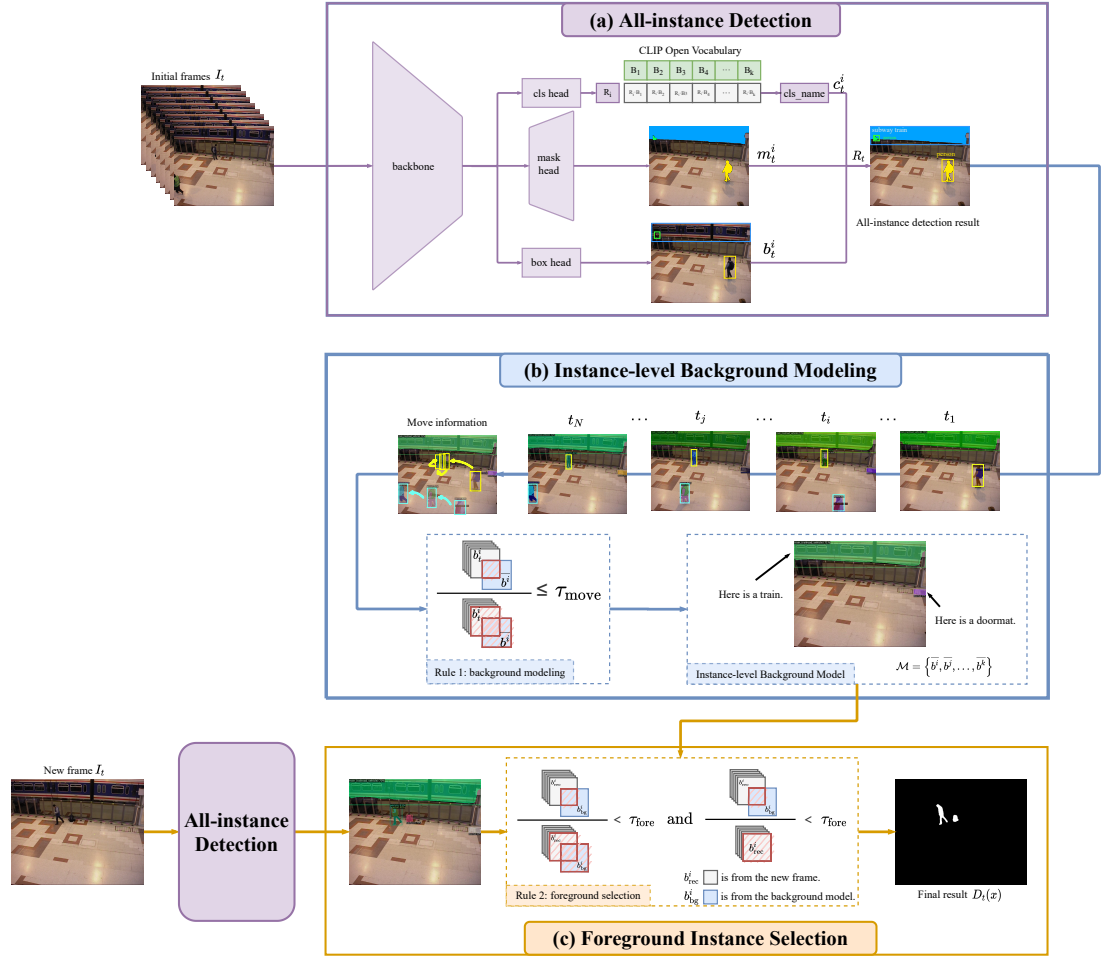} 
\caption{The framework of {\ourmodel}. (a) All-instance detection. We use a zero-shot object detection model named Detic~\cite{detic} to transform the pixel-level image into a structured-instance representation, including categories, boxes, and masks. Specifically, the categories are obtained by CLIP. (b) Instance-level background modeling. The proposed method analyzes the motion information of instances. If the instance complies with Rule 1, the stationary instance will be added to the background model. (c) The new frame output by Detic will be compared with the instance-level background model. If the instance complies with Rule 2, it will be the foreground in the final result.}
\label{fig:arch}
\end{figure*}

Over the years, deep learning-based BGS algorithms have been proposed, including supervised BGS and unsupervised BGS. Supervised BGS algorithms have achieved satisfactory performance on CDnet 2014 benchmark~\cite{FgSegNet, cascadecnn, SFEN, 3DNet, joint}. However, these methods usually have to be trained on the first several frames of the test videos, which limits the application to unseen scenarios. Unsupervised algorithms overcome this shortcoming. Most of them combine semantic segmentation models into traditional BGS algorithms. These algorithms pre-select 12 categories as foreground from 150 categories of semantic segmentation models~\cite{SemanticBGS}. Existing state-of-the-art unsupervised methods still detect night light and heavy shadows as foreground objects. As shown in \Cref{fig:pix_ins}, it is difficult for pixel-level background model to accurately distinguish the edges of foreground objects.


To tackle the above problems, we propose a novel background subtraction framework based on zero-shot object detection (ZBS). The zero-shot object detection, or also named open-vocabulary object detection, aims to detect unseen objects outside of the pre-defined categories~\cite{detic}. \Cref{fig:arch} shows the framework of our method. The method includes all-instance detection, instance-level background modeling, and foreground instance selection. In the all-instance detection stage, any zero-shot detector can be used. We use a zero-shot object detection model named Detic~\cite{detic} as the all-instance detector to transform the raw image pixels into structured instance representations, including categories, boxes, and masks. In the background modeling stage, our method builds an instance-level background model based on the motion information of instances. If an object is stationary, our algorithm adds it to the background model. In the foreground instance selection stage, the proposed algorithm selects the output of the all-instance detector when the new frame comes. If the instance complies with Rule 2 in \Cref{fig:arch} (c), it is the foreground in the final binary mask. Benefiting from the full use of instance information, our instance-level BGS method performs better in complex scenarios, such as shadows, camera jitter, night scenes, \etc. {\ourmodel} rarely detects noisy background as foreground objects by mistake. Due to the characteristics of the detector, the proposed method can detect most of the categories in the real world and can detect the unseen foreground categories outside the pre-defined categories. {\ourmodel} achieves remarkably 4.70$\%$ F-Measure improvements over state-of-the-art unsupervised methods.

Our main contributions are listed as follows:
\begin{itemize}
    \setlength{\itemsep}{0pt}
    \setlength{\topsep}{0pt}
    \setlength{\parsep}{0pt}
    \item We propose a novel background subtraction framework that has the instance-level background model;
    \item The proposed framework uses a zero-shot object detection model to obtain a more general and generalized deep learning-based unsupervised BGS algorithm; 
    \item Our method achieves the state-of-the-art in all unsupervised BGS methods on the CDnet 2014 dataset.
\end{itemize}

\section{Related work}
\label{sec:related}


\subsection{Deep learning-based Supervised Methods}
\label{subsec:deep}


Deep learning methods have been widely used for BGS due to their ability to learn high-level representations from training data~\cite{review_dl}. Braham \etal~\cite{first_deep_learning_bgs} presented the first work using deep learning for background subtraction. FgSegNet~\cite{FgSegNet} is a representative work that focuses on learning multi-scale features for foreground segmentation. CascadeCNN~\cite{cascadecnn} employs a cascade structure to synthesize the basic CNN model and the multi-scale CNN model. Zhao \etal~\cite{joint} propose an end-to-end two-stage deep CNN to reconstruct the background and separate the foreground from the background jointly. Chen \etal~\cite{SFEN} and Sakkos \etal~\cite{3DNet} use ConvLSTM and 3DCNN, respectively, to process spatio-temporal information. In addition, Siamese neural networks~\cite{cosimnet, SEU-Net}, generative adversarial networks (GAN)~\cite{BSCGAN, BSPVGAN, DBSGen}, and autoencoders (AE)~\cite{autoencoder} have also been employed for BGS.

Recently,~\cite{BSUV-Net, BSUV-Netv2, STPNet, ADNN} demonstrated better generality for unseen videos with training on limited data. However, these models are trained only on datasets containing a few categories and scenes, limiting their ability to cope with more complex real-world detection and segmentation tasks.

\subsection{Semantic background subtraction}
\label{subsec:sbs}

SemanticBGS~\cite{SemanticBGS} is the first motion detection framework to utilize object-level semantics for improving background subtraction. By combining semantic segmentation and background subtraction algorithms, it significantly reduces false positive detections and effectively identifies camouflaged foreground objects. RTSS~\cite{RTSS} performs foreground detection and semantic segmentation in a parallel manner, using the semantic probability of pixels to guide the construction and update of the background model. This method achieves real-time semantic background subtraction. RT-SBS~\cite{RT-SBS} adopts a similar approach and improves the performance, achieving a real-time semantic background subtraction algorithm at 25 frames per second.

Despite their advancements, semantic background subtraction methods are still fundamentally pixel-level background models. All of these semantic BGS methods necessitate a predefined list of foreground classes, which require expert knowledge and pose challenges for implementation in various scenarios. Furthermore, the limited number of categories in semantic segmentation networks (up to 150 categories) hinders their ability to detect moving foregrounds in an open-vocabulary setting, an aspect that is becoming increasingly important in today's environment.

The proposed {\ourmodel} builds an instance-level background model capable of detecting most real-world categories without the need for predefined foreground classes, thus offering greater practicality.

\subsection{Zero-shot object detection}
\label{subsec:zsod}
In the era of deep learning, supervised learning has achieved outstanding performance on many tasks, but too much training data has to be used. Moreover, these methods can not classify or detect the objects of categories outside the training datasets' annotations. To solve this problem, zero-shot learning~\cite{AnkanBansal2018ZeroShotOD} was proposed, hoping to classify images of categories that were never seen in the training process. The zero-shot object detection developed from this aims to detect objects outside the training vocabulary. Earlier work studied exploiting attributes to encode categories as vectors of binary attributes and learning label embeddings~\cite{AnkanBansal2018ZeroShotOD}. The primary solution in Deep Learning is to replace the last classification layer with a language embedding of class names (e.g., GloVe~\cite{Glove}). Rahman \etal~\cite{ShafinRahman2020ImprovedVA} and Li \etal~\cite{ZhihuiLi2019ZeroShotOD} improve by introducing external textual information Classifier Embeddings. ViLD~\cite{ViLD} upgrades language embeddings to CLIP~\cite{CLIP} and extracts regional features from CLIP image features. Detic~\cite{detic} also adopts CLIP as a classifier and uses additional image-level data for joint training, dramatically expanding the number of categories and performance of detection-level tasks. This paper uses the Detic detector for the BGS algorithm.

\section{Method}
\label{sec:method}

\subsection{Overview}
\label{subsec:overview}

{\ourmodel} is among the novel unsupervised BGS algorithms for real applications. It is a zero-shot object detection based on the model that can obtain an instance-level background model. {\ourmodel} contains three stages: all-instance detection, instance-level background modeling, and foreground instance selection. \Cref{fig:arch} illustrates the framework of our method. First, {\ourmodel} uses an all-instance detection model to acquire the structured-instance representation. Then, an instance-level background model is built and maintained through the movement information of instances. Finally, when a new frame comes, we will select the moving foreground from the detector outputs based on the background model. We convert the result into a binary mask to compare with other BGS methods.

\begin{algorithm}[htbp]
\caption{: The {\ourmodel} algorithm process.}
    \begin{algorithmic}[1]
        \STATE Initialize the zero-shot detector as $\mathcal{Z}$
        \STATE Initialize the background model as $\mathcal{M}$
        \STATE \textbf{while} \textit{current frame $I_t$ is valid} \textbf{do} 
        \STATE \textbf{Stage 1: All-instance detection}
        \STATE \qquad output the result $R_t$ $\leftarrow$ $\mathcal{Z}\left( I_t\right)$
        \STATE \textbf{Stage 2: Instance-level background model} \\
        \STATE \qquad get the track of each instance from $b_t$ (part of $R_t$) \\
        \STATE \qquad calculate the IoU$_{min}$ of $b^i_0 \cdots b^i_t$ and $\overline{b^i}$ \\
        \STATE \qquad update $\mathcal{M}$ based on IoU$_{min}$ and $\tau_\text{move}$ \\
        \STATE \textbf{Stage 3: Foreground instance selection} \\
        \STATE \qquad separate $\mathcal{M}$ and $b_t$ by instance-id \\
        \STATE \qquad calculate the IoU and IoF of $\mathcal{M}$ and $b^i_t$ \\
        \STATE \qquad get a binary mask $D_t(x)$ based on IoU$\&$IoF and $\tau_\text{fore}$ \\
        \STATE \qquad {current frame $\leftarrow$  next frame} \\
        \textbf{end}\\
    \end{algorithmic}
    \label{Alg:process}
\end{algorithm}

\subsection{All-instance Detection}
\label{subsec:all-id}

The goal of background subtraction is to extract all moving objects as foreground in each video frame. Traditional unsupervised BGS methods rely on pixel-level background models, which struggle to differentiate noisy backgrounds from foreground objects. To address this, we propose an instance-level background model. It utilizes an instance detector to locate the objects of all possible categories and all locations in the image and convert the raw image pixels into structured instance representation.

Intuitively, most existing trained instance segmentation networks can be used. Besides, the categories of the training datasets adapt to most domain-adapted object detection scenarios. However, instance segmentation networks cannot detect and segment the objects of categories outside the training datasets' annotations.

Recently, with the development of self-supervised training and the foundation models~\cite{CLIP}, several practical zero-shot object detection methods have been proposed~\cite{detic,ViLD}. These methods can detect almost thousands of categories of objects without being trained on the applied scenarios. Therefore, to obtain a more general and generalized deep learning background modeling method, we adopt the zero-shot object detection method Detic~\cite{detic} as the detector of our BGS method. Detic~\cite{detic} can detect 21k categories of objects and segment the object's masks.

Distinguished from the instance segmentation, we call this process \emph{all-instance detection}. After the all-instance detection stage, the video frame $I_t$ is structured by zero-shot detector $\mathcal{Z}$ as instance representation $R_t$. The representation $R_t$ includes instance boxes $b^i_t$ and segmentation masks $m^i_t$ with category labels $c^i_t$, where $i$ is the $id$ of the instance, and $t$ refers to the $t$-th frame of the video.

\subsection{Instance-level Background Modeling}
\label{subsec:instance-bm}

Based on the all-instance detection results, the algorithm should distinguish which objects have moved and which have not. The ideal instance-level background model should be the collection of all stationary object instances. It is the basis for foreground instance selection. We define the background model $\mathcal{M}$ as:
\begin{equation}
\mathcal{M}=\left\{\overline{b^i}, \overline{b^j}, \ldots, \overline{b^k}\right\}
\end{equation}

The instance-level background model $\mathcal{M}$ is a collection of detection boxes for static instances, $\overline{b^i},$ $\overline{b^j},$ $\overline{b^k}$ are static instances with $id=i, j, k$, and the value is the average of the coordinates of all boxes in the past trajectory of this instance. It reflects which locations have stationary instances and which are background without instances. As shown in \Cref{fig:arch} (b), our method uses the initial frames to obtain an initial instance-level background model and update the background model with a certain period in subsequent frames ($\Delta T=100$ is chosen in this paper). 

The details are shown in \Cref{Alg:process}. There are three steps for the instance-level background modeling stage. First, the proposed method utilizes the detector $\mathcal{Z}$ output boxes $b^i_t$ from past frames to obtain the tracks of each instance (tracks are obtained by SORT method~\cite{sort}). Second, {\ourmodel} computes the average value of the coordinates for the upper-left and lower-right corners of each bounding box within the corresponding trajectory of the instance, denoted as $\overline{b^i}$. Then we can obtain the minimum value of IoU of $b^i_t$ and $\overline{b^i}$, which means the maximum movement between the positions compared to the average in the whole trajectory. In our implementation, we apply a median filter to the trajectory IoU. This helps mitigate abrupt changes in IoU caused by object occlusion. Experiments in \Cref{tab:ablation} demonstrate that this improvement is beneficial. Finally, the update strategy of the instance-level background model is as \Cref{eq:bg_model}:
\begin{equation}\label{eq:bg_model}
\mathcal{M}= 
    \begin{cases}
        \mathcal{M} \cup \overline{b^i} \text{,} \quad \text {if  } \text{IoU}_{\min }(b_t^i, \overline{b^i}) \geqslant \tau_{\text {move}}\\ 
        \mathcal{M} - (\mathcal{M} \cap \overline{b^i}) \text{,} \quad  \text {otherwise.}
    \end{cases}
\end{equation}

\noindent{where $b^i_t$ denotes the $i$-th instance in $t$-th image frame. $\overline{b^i}$ denotes the average of all boxes $\overline{b^i}$ for each instance. $\tau_\text{move}$ is the threshold for judging whether the instance is moving. If it remains stationary, put it into the background model $\mathcal{M}$; otherwise, remove it from the background model $\mathcal{M}$.}

\noindent{\textbf{Implementation Details.} To build a more robust background model, we choose a smaller $\tau_\text{conf}$\footnote{$\tau_\text{conf}$ is a score threshold from~\cite{detic} for the outputs of the all-instance detection stage. This threshold determines the confidence level of the outputs in the first stage.} in the instance-level background modeling stage, which helps more stationary instances to be incorporated into the background model $\mathcal{M}$, it is called $\Delta$conf.


\subsection{Foreground Instance Selection}
\label{subsec:fis}

The key to foreground instance selection is accurately judging whether the instance has moved compared to the background model. Object occlusion is a common challenge. When objects are occluded, the object behind them can easily be misjudged as moving. To balance sensitivity and robustness to object occlusion, we introduce IoF (Intersection over Foreground) as a complement to IoU (Intersection over Union) , which is calculated as \Cref{eq:iof}:
\begin{equation}\label{eq:iof}
    \text{IoF}=\frac{b_\text{rec}^i \cap b_\text{bg}^i}{b_\text{rec}^i}.
\end{equation}

\noindent{where $b_\text{rec}^i$ denotes the instance in the recent frame, $b_\text{bg}^i$ denotes the $i$-th instance in the instance-level background model $\mathcal{M}$.}

\begin{figure}[ht]
\centering
\includegraphics[width=0.9\columnwidth]{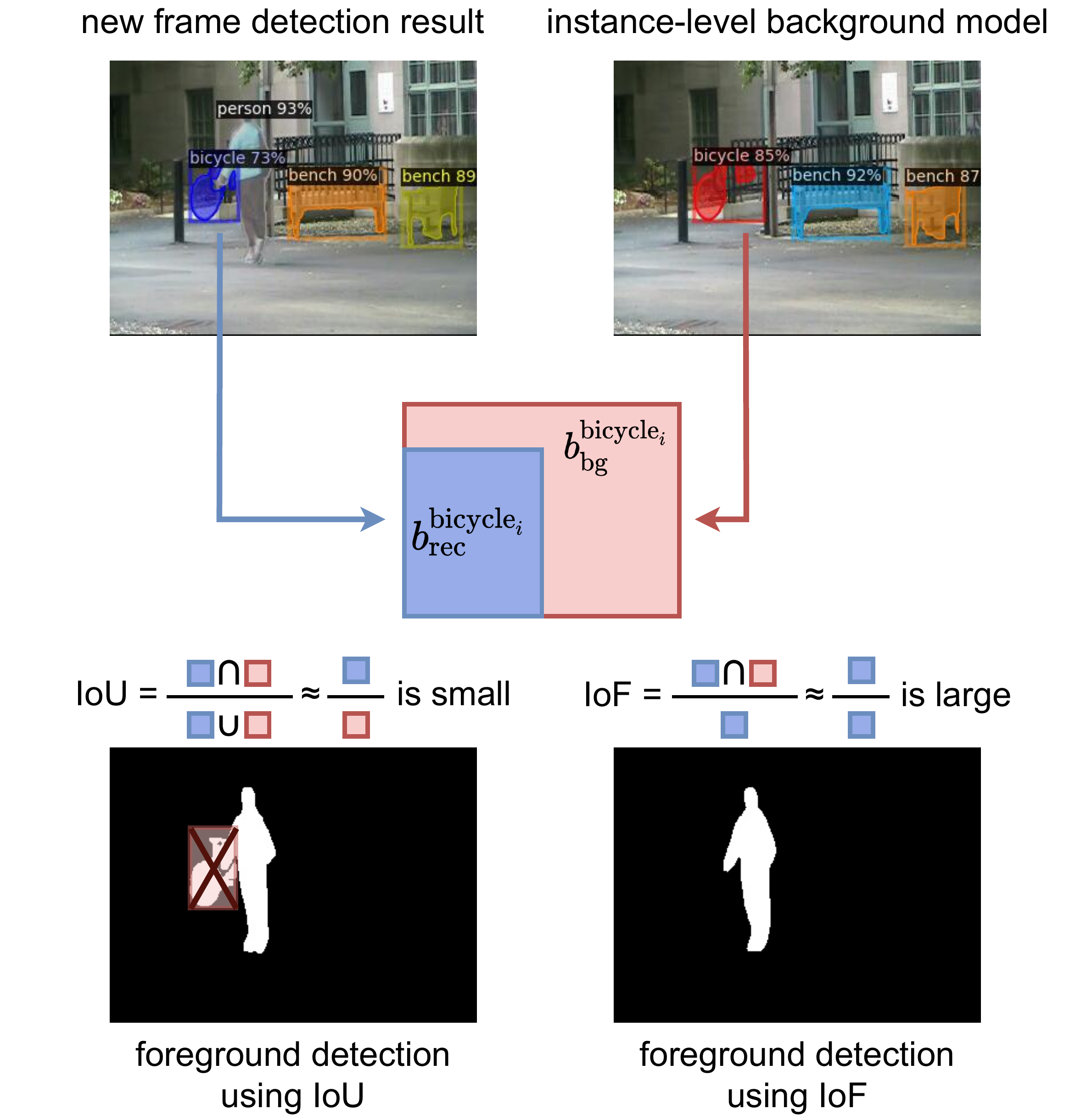} 
\caption{A typical case for object obstruction. A person half obscures the bicycle on the left side of the image. The IoU of $b_{\text {rec}}^{\text {bicycle}_i}$ and $b_{\text {bg}}^{\text {bicycle}_i}$ is very small, while the IoF of $b_{\text {rec}}^{\text {bicycle}_i}$ and $b_{\text {bg}}^{\text {bicycle}_i}$ is still large.}
\label{fig:iou_iof}
\vspace{-0.4cm}
\end{figure}

\Cref{fig:iou_iof} shows how IoF works. If the instance is not moving but is obscured, IoU drops considerably while IoF still preserves a high value. By judging the two metrics together, it is robust to determine whether a certain instance is a foreground.


As shown in the foreground instance selection stage of \Cref{Alg:process}, the detection result of each current frame is matched with the set of object instances contained in the instance-level background model. The proposed method uses two metrics, IoU and IoF, to determine whether the instance can be used as a foreground. If both IoU and IoF are smaller than the foreground selection threshold $\tau_\text{fore}$, the instance is new or has moved and should be considered as foreground. On the contrary, if either IoU or IoF is larger than the threshold $\tau_\text{fore}$, the instance should not be considered as foreground. The rule of foreground instance selection can be expressed by \Cref{eq:fore}.
\begin{equation}\label{eq:fore}
\begin{split}
D_t(x)= \left \{
\begin{array}{ll}
    FG \text{, if IoU}(b_\text{rec}^i, b_\text{bg}^i) < \tau_\text{fore} \\
    \qquad \text{and IoF}(b_\text{rec}^i, b_\text{bg}^i) < \tau_\text{fore};\\
    BG \text{, otherwise}.
\end{array}
\right.
\end{split}
\end{equation}

\noindent{where $D_t(x)$ is regarded as the $x$-th instance in $t$-th frame whether should be a foreground, $FG$ means foreground, $BG$ means background. $b_\text{rec}^i$ denotes the box of $i$-th instance in recent frame. $b_\text{bg}^i$ denotes the box of foreground instance in the instance-level background model $\mathcal{M}$.}

\section{Experiments}
\label{sec:exp}

\subsection{Dataset and Evaluation Metrics}
\label{subsec:data-eval}

We evaluate the performance of the proposed method on the CDnet 2014 dataset~\cite{CDnet2014}. The CDnet 2014 dataset is the most famous benchmark for change detection, including 53 video sequences and 11 categories corresponding to different classic application scenarios. The main categories of the dataset include \emph{Bad Weather}, \emph{Baseline}, \emph{Camera Jitter}, \emph{Dynamic Background}, \emph{Intermittent Object Motion}, \emph{Low Framerate}, \emph{Night videos}, \emph{Pan–Tilt–Zoom}, \emph{Shadow}, \emph{Thermal}, and \emph{Turbulence}. Ground Truth annotated manually is available for every frame in video sequences and tells us whether each pixel belongs to the background or the foreground. Specifically, in the Ground Truth, Static, Shadow, Non-Region of Interest (Non-ROI), Unknown, and Moving pixels are assigned respectively to grayscale values 0, 50, 85, 170, and 255. We select Recall (\emph{Re}), Precision (\emph{Pr}) and F-Measure (\emph{F$-$M}) to evaluate the performance on the CDnet 2014 dataset.


Following~\cite{cdnet2012}, we regard Static and Shadow pixels as negative samples (background), regard Moving pixels as positive samples (foreground), and ignore Non-ROI and Unknown pixels to ensure the fairness of the metrics.

\begin{figure*}[t]
    \centering
    \begin{subfigure}{0.33\linewidth}
        \includegraphics[width=1\linewidth]{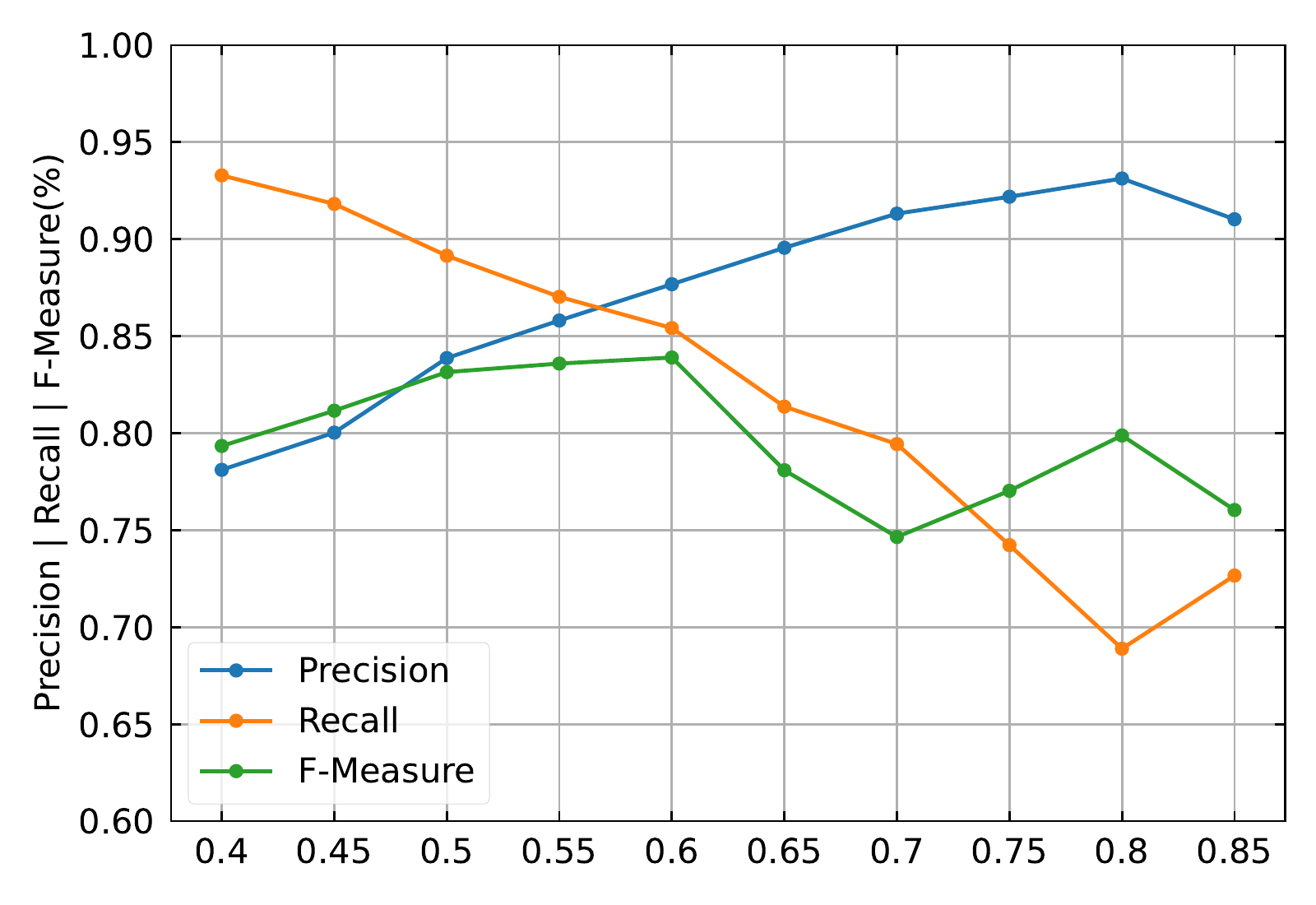}
        \caption{The sensitivity of the threshold $\tau_\text{conf}$}
    \end{subfigure}
    \hfill
    \begin{subfigure}{0.33\linewidth}
        \includegraphics[width=1\linewidth]{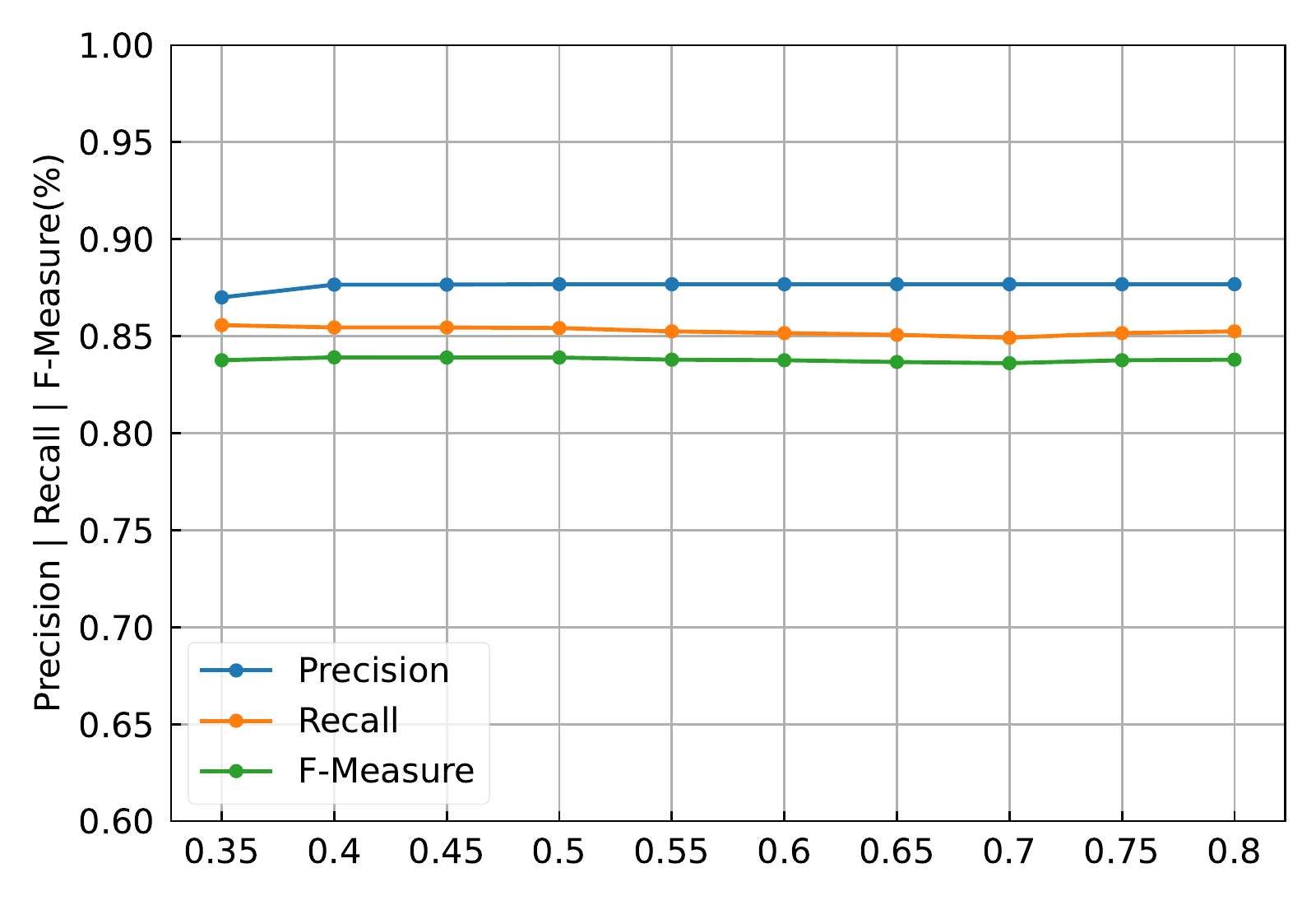}
        \caption{The sensitivity of the threshold $\tau_\text{move}$}
    \end{subfigure}
    \begin{subfigure}{0.33\linewidth}
        \includegraphics[width=1\linewidth]{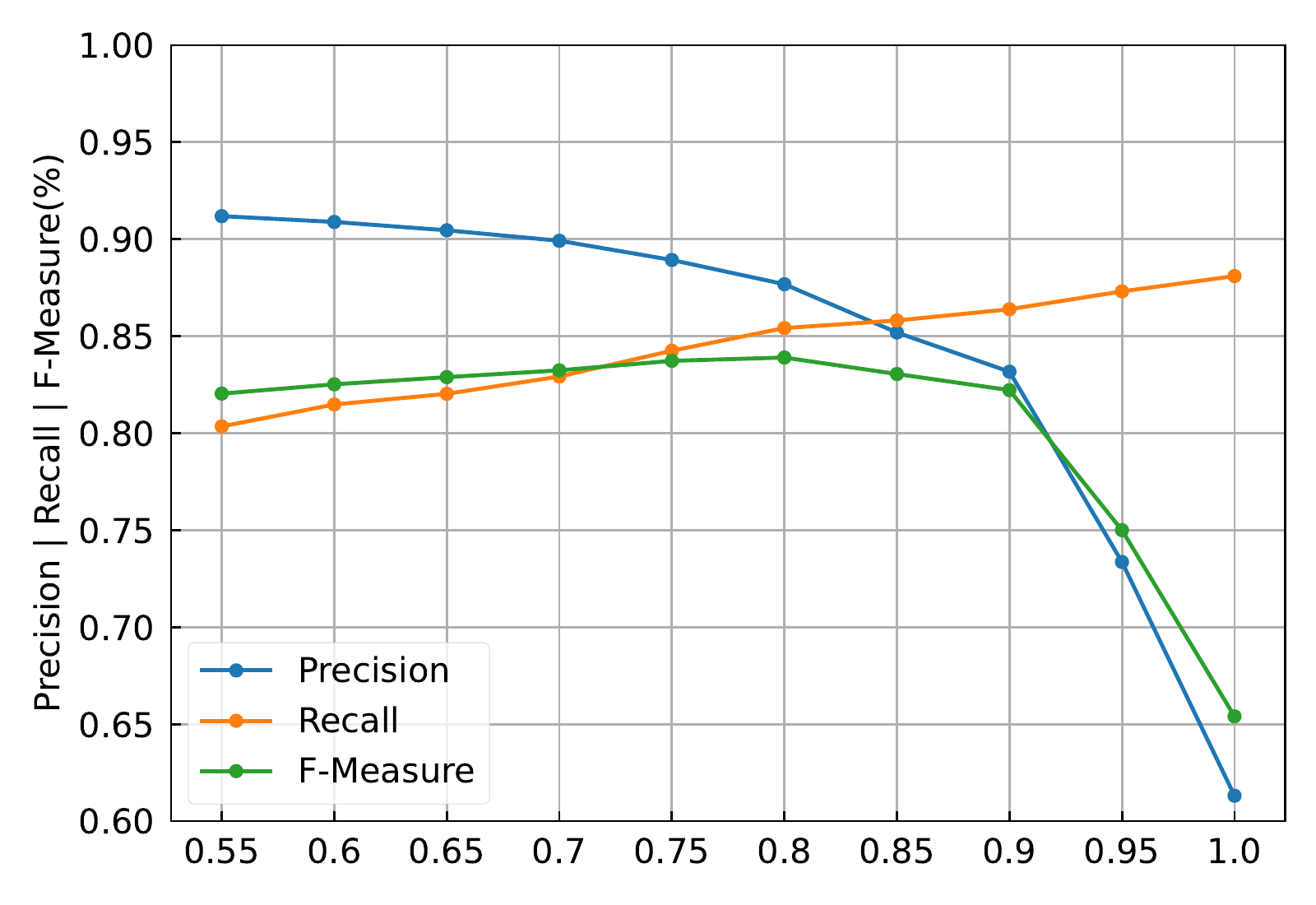}
        \caption{The sensitivity of the threshold $\tau_\text{fore}$}
    \end{subfigure}
    \caption{The hyper-parameter sensitivity analysis. The relationship between the thresholds ($\tau_\text{conf},$ $\tau_\text{move},$ $\tau_\text{fore}$) and the evaluation metrics (Precision, Recall, F-Measure) of the BGS algorithms.}
    \label{fig:sensitivity}
    \vspace{-0.2cm}
\end{figure*}

\subsection{Hyper-parameter Sensitivity Analysis}
\label{subsec:param}

As mentioned earlier, our method requires setting several parameters before use: the threshold for all-instance detection $\tau_\text{conf}$, the threshold for background modeling $\tau_\text{move}$, and the threshold for foreground selection $\tau_\text{fore}$. 

Different parameters suit different situations. The large $\tau_\text{conf}$ is better for more straightforward scenarios. The large $\tau_\text{move}$ is better for fast motion or low frame rate scenarios. The small $\tau_\text{fore}$ is robust for camera jitter scenarios. The performance with different parameter settings is shown in \Cref{fig:sensitivity}. For most scenarios, the sensitivity of the two parameters $\tau_\text{move},$ $\tau_\text{fore}$ is low, and the impact of the changes of these two hyper-parameters on 
F-Measure fluctuates within ±1$\%$. When $\tau_\text{fore}$ is equal to 1, the foreground instance selector treats all new instances as foreground, so the precision and F-Measure have a big drop. $\tau_\text{conf}$ determines the output of the zero-shot detector, which is very important for the subsequent two stages, and different scenes often apply to different thresholds. The universal parameters are $\tau_\text{conf}=0.6,$ $\tau_\text{move}=0.5,$ $\tau_\text{fore}=0.8$ in Experiments.


\subsection{Quantitative Results}
\label{subsec:quant}

\Cref{tab:compare-sota} shows the comparison of our method among other BGS algorithms, in which F-Measure is observed. These algorithms could be classified into two parts: supervised and unsupervised algorithms. Most supervised algorithms, such as FgSegNet~\cite{FgSegNet} and CascadeCNN~\cite{cascadecnn}, have nearly perfect F-Measure because they are trained with some frames in test videos. However, the performance of these methods decreases significantly when applied to unseen videos because of the lack of generalization ability. FgSegNet in unseen videos only achieves 0.3715 F-Measure. STPNet~\cite{STPNet} and BSUV-Net 2.0~\cite{BSUV-Netv2} are supervised algorithms designed explicitly for unseen videos and can achieve F-Measure of around 0.8. IUTIS-5~\cite{IUTIS} is a special supervised algorithm that learns how to combine various unsupervised algorithms from datasets.

The remaining methods are unsupervised algorithms~\cite{SWCD, Wisenetmd, SuBSENSE, PAWCS, RTSS} which naturally can handle unseen videos. The results show that our method outperforms all unsupervised algorithms. In particular, ZBS outperforms the state-of-the-art RT-SBS-v2~\cite{RT-SBS} by 4.70$\%$. Moreover, ZBS outperforms supervised method in unseen videos BSUV-Net 2.0~\cite{BSUV-Netv2}. When considering per-category F-Measure, our method has advantages in seven out of eleven categories, such as \emph{Camera Jitter}, \emph{Intermittent Object Motion}, and \emph{Night Videos}. However, our method cannot deal with \emph{Turbulence} well because the detector of the all-instance detection module cannot adapt to the unnatural image distribution of \emph{Turbulence} scenarios without training.

\begin{table*}[ht]
    \begin{center}
    \caption{Overall and per-category F-Measure comparison of different BGS methods on the CDnet 2014 dataset.}
    \vspace{-0.3cm}
    \tabcolsep=5pt
    \smallskip
    \scalebox{0.85}{
        \begin{tabular}{c|ccccccccccc|c}
            \toprule
            Method  & baseline  & camjitt   & dynbg  & intmot & shadow  & thermal   & badwea  & lowfr   & night  & PTZ   & turbul  & Overall  \\
            \midrule
            \multicolumn{13}{c}{Supervised algorithms}\\
            \midrule
            {CascadeCNN~\cite{cascadecnn}}&   {0.9786}&   {0.9758}&   {0.9658}&   {0.8505}&   {0.9593}&   {0.8958}&   {0.9431}&   {0.8370}&   {0.8965}&   {0.9168}&   {0.9108}&   {0.9209}\\
            {MU-Net2~\cite{MU-Net2}}&   {0.9900} &  {0.9824}&   {0.9892}&   {0.9894}&   {0.9845}&   {0.9842}&   {0.9343}&   {0.8706}&   {0.8362}&   {0.8185}&   {0.9272}&   {0.9369}\\
            {BSPVGAN~\cite{BSPVGAN}}&   {0.9837} &  {0.9893}&   {0.9849}&   {0.9366}&   {0.9849}&   {0.9764}&   {0.9644}&   {0.8508}&   {0.9001}&   {0.9486}&   {0.9310}&   {0.9501}\\
            {FgSegNetv2~\cite{FgSegNetv2}}&   {0.9978} &  {0.9971}&   {0.9951}&   {0.9961}&   {0.9955}&   {0.9938}&   {0.9904}&   {0.9336}&   {0.9739}&   {0.9862}&   {0.9727}&   {0.9847}\\
            \makecell[c]{FgSegNet~\cite{FgSegNet}\\(unseen video)}&   {0.6926} &  {0.4266}&   {0.3634}&   {0.2002}&   {0.5295}&   {0.6038}&   {0.3277}&   {0.2482}&   {0.2800}&   {0.3503}&   {0.0643}&   {0.3715}\\
            {STPNet~\cite{STPNet}}&   {0.9587}&   {0.7721}&   {0.8058}&   {0.8267}&   {0.9114}&   {0.8688}&   {0.8898}&   {0.7297}&   {0.6961}&   {0.6076}&   {0.7248}&   {0.7992}\\
            {BSUV-Net 2.0~\cite{BSUV-Netv2}}&   {0.9620}&   {0.9004}&   {0.9057}&   {0.8263}&   {0.9562}&   {0.8932}&   {0.8844}&   {0.7902}&   {0.5857}&   {0.7037}&   {0.8174}&   {0.8387}\\
            {IUTIS-5~\cite{IUTIS}}&   {0.9567}&   {0.8332}&   {0.8902}&   {0.7296}&   {0.8766}&   {0.8303}&   {0.8248}&   {0.7743}&   {0.5290}&   {0.4282}&   {0.7836}&   {0.7717}\\
            \midrule
            \multicolumn{13}{c}{Unsupervised algorithms}\\
            \midrule
            {PAWCS~\cite{PAWCS}}&   {0.9397}&   {0.8137}&   {0.8938}&   {0.7764}&   {0.8913}&   {0.8324}&   {0.8152}&   {0.6588}&   {0.4152}&   {0.4615}&   {0.6450}&   {0.7403}\\
            {SuBSENSE~\cite{SuBSENSE}}&   {0.9503}&   {0.8152}&   {0.8177}&   {0.6569}&   {0.8986}&   {0.8171}&   {0.8619}&   {0.6445}&   {0.5599}&   {0.3476}&   {0.7792}&   {0.7408}\\
            {WisenetMD~\cite{Wisenetmd}}&   {0.9487}&   {0.8228}&   {0.8376}&   {0.7264}&   {0.8984}&   {0.8152}&   {0.8616}&   {0.6404}&   {0.5701}&   {0.3367}&   \textbf{0.8304}&   {0.7535}\\
            {SWCD~\cite{SWCD}}&   {0.9214}&   {0.7411}&   {0.8645}&   {0.7092}&   {0.8779}&   {0.8581}&   {0.8233}&   {0.7374}&   {0.5807}&   {0.4545}&   {0.7735}&   {0.7583}\\
            {SemanticBGS~\cite{SemanticBGS}}&   {0.9604}&   {0.8388}&  \textbf{0.9489}&   {0.7878}&   {0.9478}&   {0.8219}&   {0.8260}&   \textbf{0.7888}&   {0.5014}&   {0.5673}&   {0.6921}&   {0.7892}\\
            {RTSS~\cite{RTSS}}&   {0.9597}&   {0.8396}&   {0.9325}&   {0.7864}&   {0.9551}&   {0.8510}&   {0.8662}&   {0.6771}&   {0.5295}&   {0.5489}&   {0.7630}&   {0.7917}\\
            {RT-SBS-v2~\cite{RT-SBS}}&   {0.9535}&   {0.8233}&  {0.9217}&   \textbf{0.8946}&   {0.9497}&   {0.8697}&   {0.8279}&   {0.7341}&   {0.5629}&   {0.5808}&   {0.7315}&   {0.8045}\\
            {{\ourmodel} (Ours)}&   \textbf{0.9653}&   \textbf{0.9545}&   {0.9290}&   {0.8758}&   \textbf{0.9765}&   \textbf{0.8698}&   \textbf{0.9229}&   {0.7433}&   \textbf{0.6800}&   \textbf{0.8133}&   {0.6358}&   \textbf{0.8515}\\
            \bottomrule
    \end{tabular}
    }
    \label{tab:compare-sota}
    \end{center}
    \vspace{-0.5cm}
\end{table*}

\subsection{Ablation Study}
\label{subsec:ablation}

Ablation experiments are conducted, in which we add the ablation components one by one to measure their effectiveness. The results are summarized in \Cref{tab:ablation} with precision, recall, and F-Measure. 

The baseline is to use the result of the all-instance detector directly as the foreground. In the instance-level background modeling stage, we only build an instance-level background model, but do not judge whether the foreground is moving. The performance of the algorithm is slightly improved and exceeds the baseline. In the foreground selection stage, the algorithm uses the background model to determine whether the foreground is moving. The performance is greatly improved. Moreover, we propose three modules to enhance the instance-level background model. After adding them to the algorithm one by one, the algorithm's performance larger gains and the advantages of the instance-level background model are fully demonstrated.

\Cref{tab:ablation} shows that simple instance-level background and foreground capture most potential moving objects. The former exhibits higher recall but slightly lower precision than pixel-level background and foreground. $\Delta$conf enhances overall performance. Object occlusion impacts the background model and complicates foreground selection. This issue is unique to instance-level representation. We propose the "IoU filter" and "IoF" to mitigate this problem, both of which reduce false positives, particularly "IoF".

\begin{table}[ht]
    \centering
    \caption{
    Ablation of the three stages of our methods and other improvements. AID: All-instance detection (\Cref{subsec:all-id}). IBM: Instance-level background modeling (\Cref{subsec:instance-bm}). FIS: Foreground instance selection (\Cref{subsec:fis}). $\Delta$conf: Different confidence thresholds for background modeling and foreground selection. Filter: Median filtering on movement information. IoF: Intersection over Foreground measurement standard.
    }
    \vspace{-0.2cm}
    \smallskip

    \scalebox{0.75}{
        \begin{tabular}{cccccc|ccc}
            \toprule
            \multirow{2}{*}{AID} & \multirow{2}{*}{IBM} & \multirow{2}{*}{FIS} & \multicolumn{3}{c|}{Enhance} & \multirow{2}{*}{Pr} & \multirow{2}{*}{Re} & \multirow{2}{*}{F-M} \\ \cline{4-6}
            & & & $\Delta$conf & Filter & IoF \\
             \midrule
              \ding{52} & & & & & & 0.4076 & 0.8869 & 0.4980\\
              \hline
              \ding{52}&\ding{52} & & & & & 0.5343 & 0.8022 & 0.5752\\
             \hline
             \ding{52}&\ding{52} & \ding{52} & & & & 0.7468 & 0.7625 & 0.7152\\
             \hline
              \ding{52}&\ding{52} &\ding{52} &\ding{52} & & & 0.7529 & 0.7851 & 0.7415\\
             \hline
             \ding{52}&\ding{52} &\ding{52} &\ding{52}&\ding{52} & & 0.8249 & 0.7829 & 0.7836\\
             \hline
             \ding{52}&\ding{52} &\ding{52} &\ding{52} &\ding{52} &\ding{52} & \textbf{0.8802} & \textbf{0.8403} & \textbf{0.8515}\\
             \bottomrule
        \end{tabular}}
        \label{tab:ablation}
\vspace{-0.8cm}
\end{table}

\subsection{More Analysis}
\label{subsec:more}

\noindent{\textbf{Visual Result.} A visual comparison from different methods is shown in \Cref{fig:visual}. It includes six challenging scenarios from the CDnet 2014 dataset. In the "highway" scenario, the main challenge is the shadow of the car and the tree branches moving with the wind. Because of the instance-level foreground detection, our method is robust to noisy background regions. In the "boulevard" scenario, affected by the jitter of the camera, many false positives are produced by other BGS methods. However, our method is robust to camera shake due to the instance-level background model. In the "boats" scenario, GMM and SuBSENSE produce many false positives because of water rippling. {\ourmodel} can extract clear detection results within the background disturbance. In the "sofa" scenario, which contains intermittent object motion, the proposed method has better segmentation results at the contour regions of objects. In the "peopleInShade" scenario, {\ourmodel} excels in shadow regions. Despite challenges in the "continuousPan" scenario, our method remains robust. These results highlight the advantages of our instance-level background-based {\ourmodel}.

\begin{figure*}[t]
\centering
\includegraphics[width=0.78\linewidth]{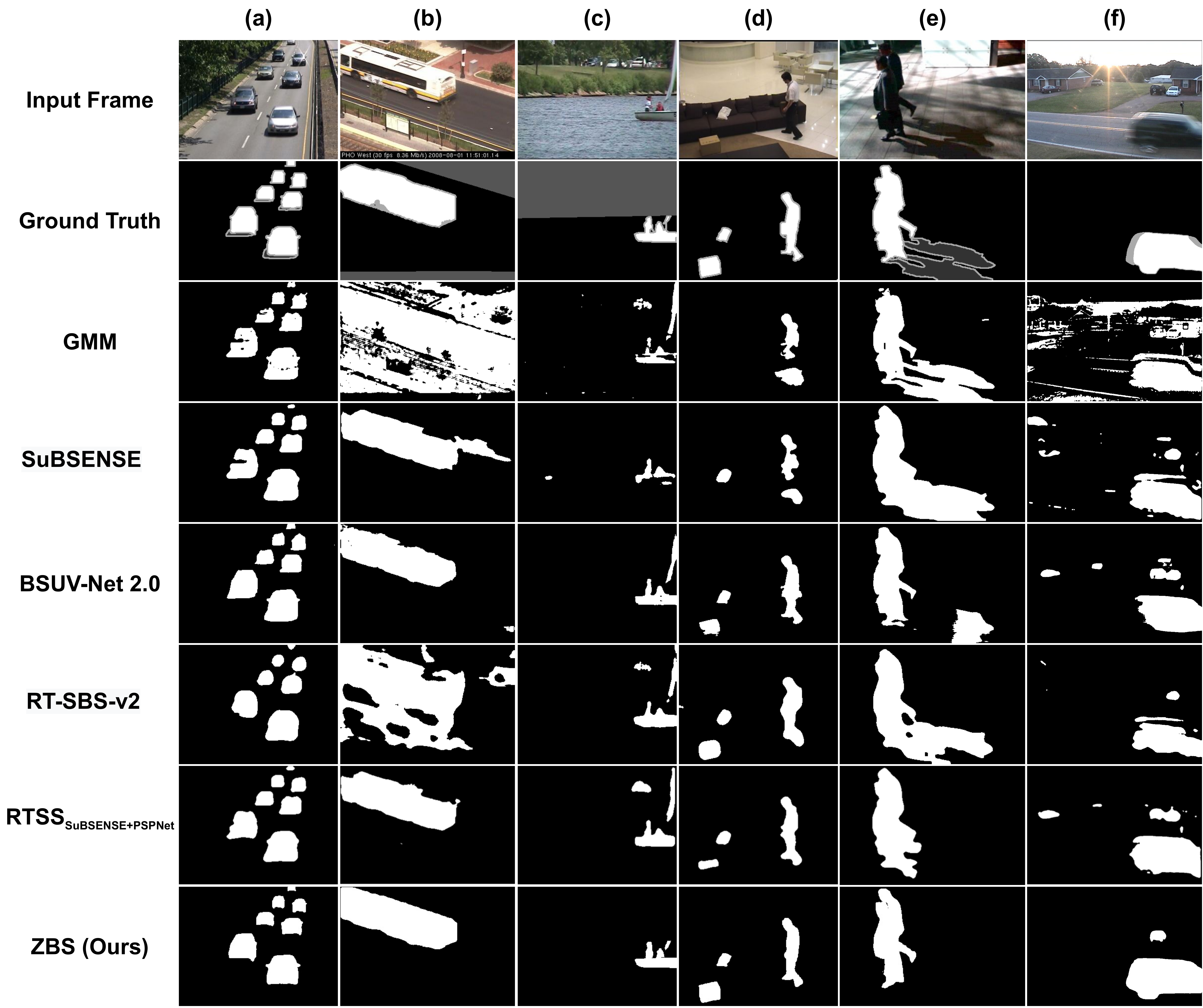} 
\caption{Comparison of the visual results on various scenarios from CDnet 2014 dataset. Except for our method, other segmentation results are quoted in RTSS~\cite{RTSS}. From left to right: (a) Scenario "highway" from the \emph{Baseline} category. (b) Scenarios "boulevard" from the \emph{Camera Jitter} category. (c) Scenario "boats" from the \emph{Dynamic Background} category. (d) Scenario "sofa" from the \emph{Intermittent Object Motion} category. (e) Scenario "peopleInShade" from the \emph{Shadow} category. (f) Scenario "continuousPan" from the \emph{PTZ} category.}
\label{fig:visual}
\vspace{-0.4cm}
\end{figure*}

\noindent{\textbf{Runtime Efficiency.} Achieving real-time performance is vital for BGS algorithms. The main time-consuming aspect of {\ourmodel} is concentrated in the first stage, which involves pre-training the zero-shot object detection model. We have implemented the two subsequent stages in C++. The FPS is about 20 on one A100 GPU. In the all-instance detection stage, we used parallel computation with a batch size of 8 and processed results sequentially in the subsequent stages. Ultimately, we achieve approximately 44 FPS on an A100 GPU.


Moreover, adopting TensorRT SDK \cite{vanholder2016efficient}, quantization, and frame skipping can further improves the FPS in real-world applications. However, this paper mainly focus on enhancing the accuracy of BGS (F-Measure). In future studies, we plan to further improve its runtime efficiency.}

\noindent{\textbf{Performance in complex scenarios.} To further demonstrate the good performance of {\ourmodel} in complex scenarios, we compare the nbShadowError and FPR-S of different BGS methods in the \emph{Shadow} category. FPR-S is calculated as \Cref{eq:fpr-s}. \Cref{tab:fpr-s} shows that our method has an extremely low false positive rate on shadow pixels. {\ourmodel} far outperforms all unsupervised BGS methods and is better than the state-of-the-art supervised method FgSegNet~\cite{FgSegNet}. In the appendix, we also add experiments to demonstrate the good performance in other complex scenes such as night light, camouflaged foreground, \etc.
\begin{equation} \label{eq:fpr-s}
    \text{FPR-S}= \text{nbShadowError}/ \text{nbShadow}
\end{equation}

\noindent{where nbShadowError is the number of times a pixel is labeled as shadow in Ground Truth but detected as a moving object. nbShadow is the total number of pixels labeled as a shadow in Ground Truth for a video or category.}

\begin{table}[ht]
\begin{center}
\caption{The FPR-S and nbShadowError of different BGS methods.}
\vspace{-0.4cm}
\tabcolsep=4pt
\smallskip
\scalebox{0.7}{
    \begin{tabular}{c|cccc|c}
    \toprule
    \multirow{2}{*}{Method} & \multicolumn{4}{c|}{nbShadowError} & \multirow{2}{*}{FPR-S} \\ \cline{2-5}
            & busStation & peopleInShade & bungalows & cubicle  \\
    \midrule
    {FgSegNet~\cite{FgSegNet}}&   {2383}&   {12866}&  \textbf{5375}&   {580}&   {0.0042}\\
    {BSUV-Net 2.0~\cite{BSUV-Netv2}}&   {23149}&   {564989}&  {982943}&   {111438}&   {0.2506}\\
    {SuBSENSE~\cite{SuBSENSE}}&   {315658}&   {854157}&  {1705793}&   {391569}&   {0.5996}\\
    {SemanticBGS~\cite{SemanticBGS}}&   {169426}&   {782489}&  {730668}&   {33137}&   {0.3018}\\
    {RT-SBS-v2~\cite{RT-SBS}}&   {28530}&   {457467}&  {566642}&   {52746}&   {0.1717}\\
    {{\ourmodel} (Ours)}&   \textbf{964}&   \textbf{1892}&  {10403}&   \textbf{390}&   \textbf{0.0019}\\
    \bottomrule
\end{tabular}}
\label{tab:fpr-s}
\end{center}
\vspace{-0.6cm}
\end{table}


\section{Conclusion}
\label{sec:conclusion}

In this paper, we propose a novel background subtraction framework, {\ourmodel}, consisting of three components: all-instance detection, instance-level background modeling, and foreground instance selection. Experiments on the CDnet 2014 dataset show the algorithm's effectiveness. Compared with other BGS methods, our method achieves state-of-the-art performance among all unsupervised BGS methods and even outperforms many supervised deep learning algorithms. {\ourmodel} detects most real-world categories without pre-defined foreground categories, producing accurate foreground edges and reducing false detections.

Our future work is leveraging instance-level information more effectively and compressing the all-instance detector for better efficiency.

\noindent{\textbf{Acknowledgement.} This work was supported by National Key R$\&$D Program of China under Grant No.2021ZD0110403. This work was also supported by National Natural Science Foundation of China under Grants 61976210, 62176254, 62006230, 62002357, and 62206290.




{\small
\bibliographystyle{ieee_fullname}
\bibliography{egbib}
}

\appendix

\section{Instance-level foreground detection}

Unlike previous methods, our method builds an instance-level background model. Therefore, {\ourmodel} can achieve instance-level foreground detection. \Cref{fig:ins_detect} shows the difference between binary foreground detection and instance-level foreground detection. \Cref{fig:ins_detect(b)} shows that {\ourmodel} can detect moving foreground of different granularities, including person, backpack, shoe, beanie, \etc., and can correctly classify stationary subway and crossbar as background.}

\vspace{-0.2cm}
\begin{figure}[ht]
    \centering
    \begin{subfigure}{0.43\linewidth}
        \includegraphics[width=1\linewidth]{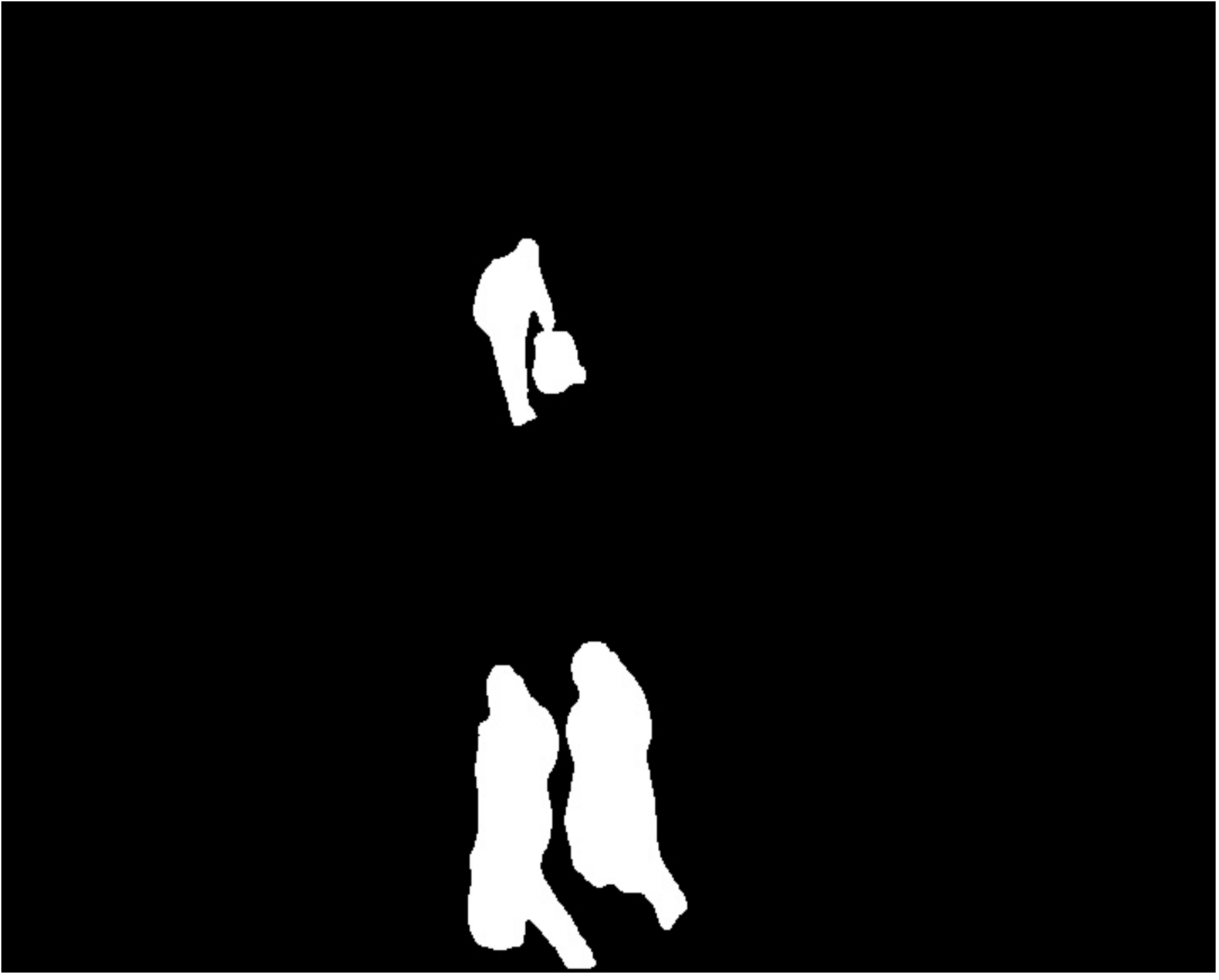}
        \caption{Binary foreground detection.}
        \label{fig:ins_detect(a)}
    \end{subfigure}
    \hfill
    \begin{subfigure}{0.43\linewidth}
        \includegraphics[width=1\linewidth]{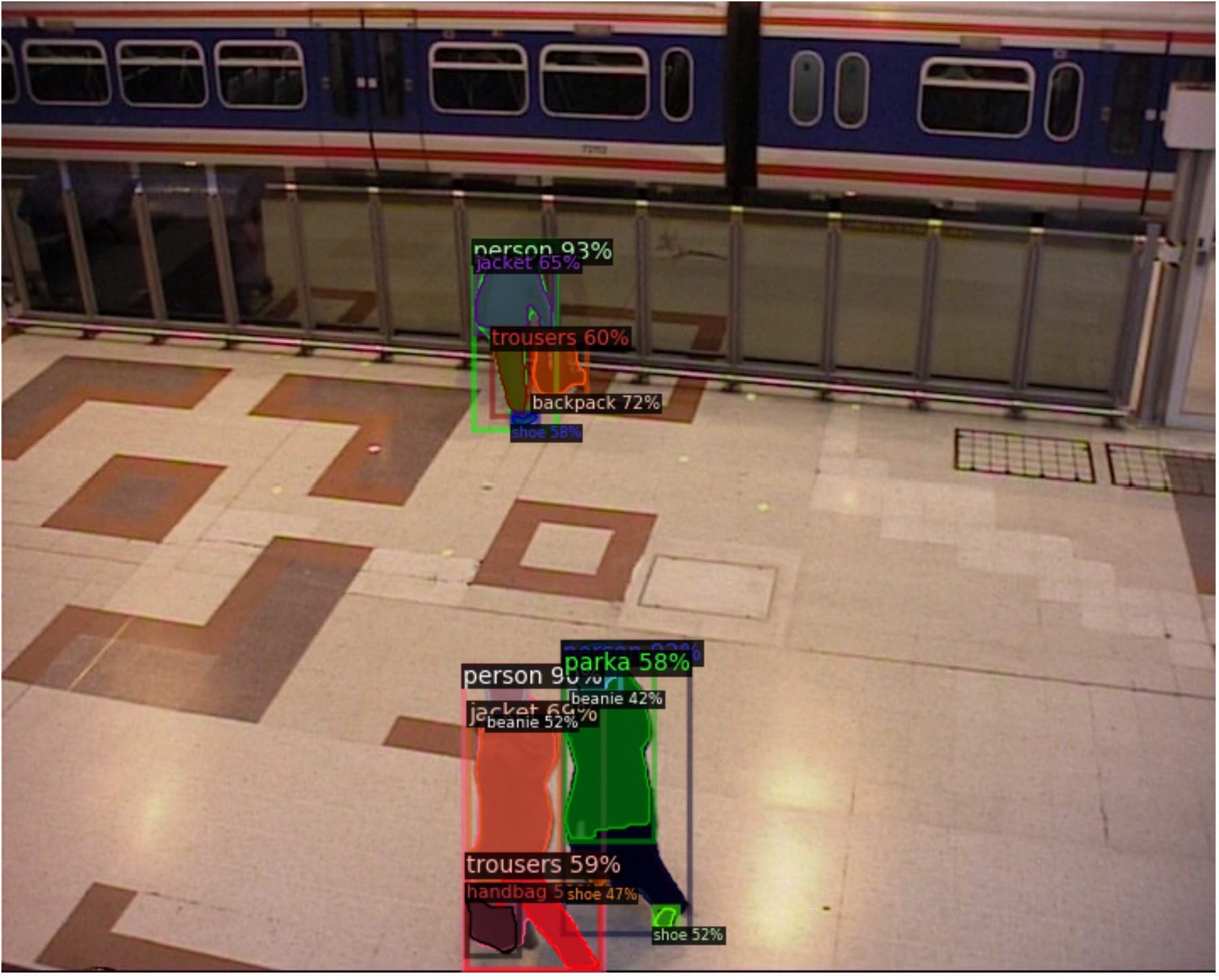}
        \caption{Instance-level foreground detection.}
        \label{fig:ins_detect(b)}
    \end{subfigure}
    \caption{The binary and instance-level foreground detection of {\ourmodel}. Our method can detect the moving foreground of different granularities.}
    \label{fig:ins_detect}
\end{figure}

\section{Abandoned object detection}

Abandoned object detection in video surveillance is critical for ensuring public safety and is a crucial component of Intelligent Monitoring. This task presents a challenge, as the categories of abandoned objects are highly diverse and difficult to learn through traditional supervised training methods. Traditional background subtraction techniques often prove insufficient in addressing this issue. Our proposed method, however, offers a solution by incorporating a stronger semantic discernment and instance-level background model, resulting in effective detection of abandoned objects.

To adapt to new tasks, we have added a new rule that considers both motion information and the relationships between instances. If an object exhibits isolated, static behavior or moves independently after previously moving in sync with categories such as a person or car, the instance is deemed to be an abandoned object. This straightforward semantic rule has proven to be effective in diverse environments. We have conducted thorough experiments on the public datasets PETS2006 and ABODA, as well as a non-public traffic abandoned object detection dataset known as TADA.

\subsection{PETS2006}

The PETS2006 dataset includes sequences from seven different scenes. Each sequence contains an abandonment event except for the third event. We evaluates all seven sequences and our method successfully detectes the abandoned objects for the entire PETS2006 dataset without any false alarms. As shown in \Cref{fig:pets}, the results from the PETS2006 dataset demonstrate the efficacy of our approach.

\vspace{-0.2cm}
\begin{figure}[ht]
    \centering
    \begin{subfigure}{0.32\linewidth}
        \includegraphics[width=1\linewidth]{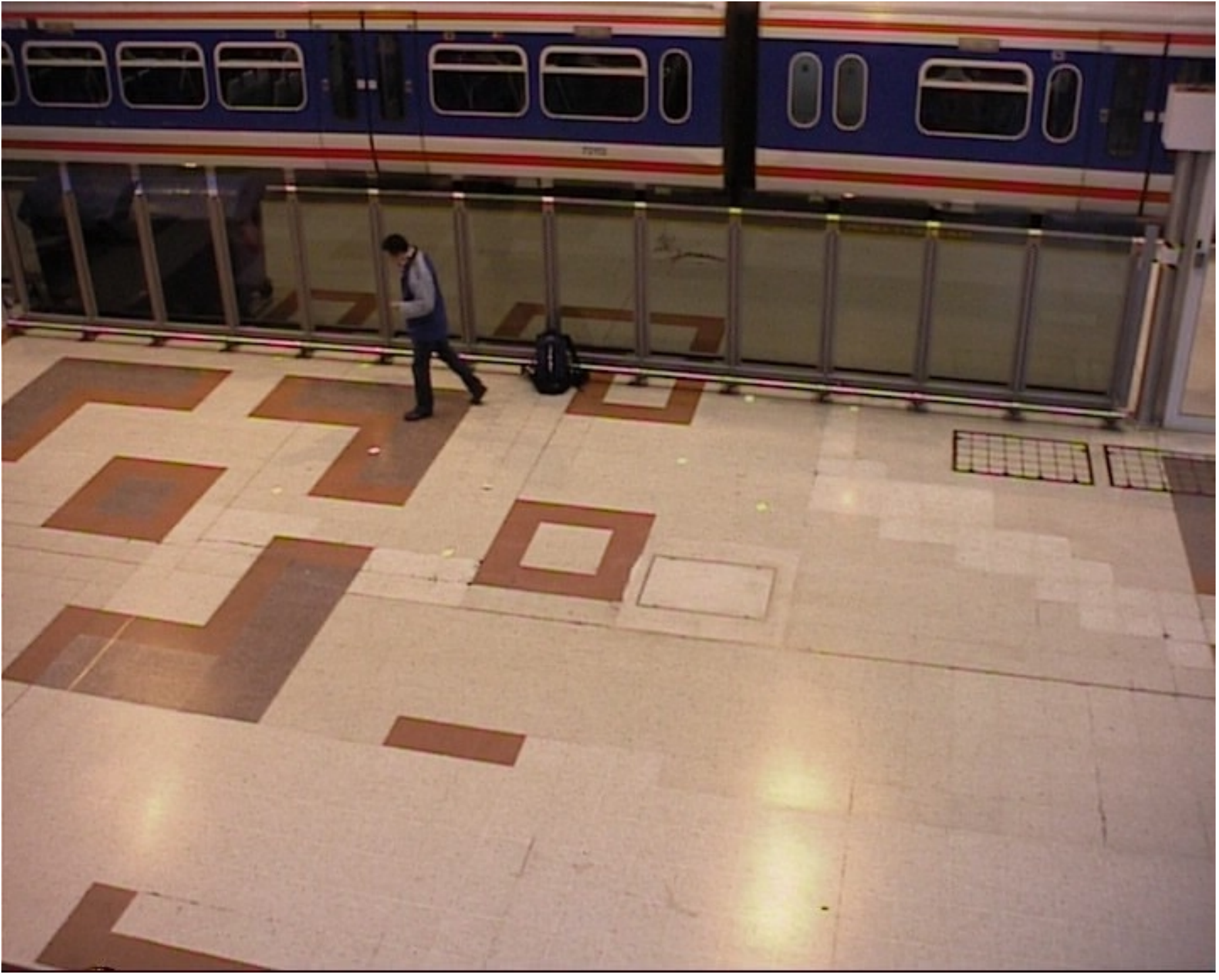}
        \caption{}
    \end{subfigure}
    \hfill
    \begin{subfigure}{0.32\linewidth}
        \hspace{-0.18cm}
        \includegraphics[width=1\linewidth]{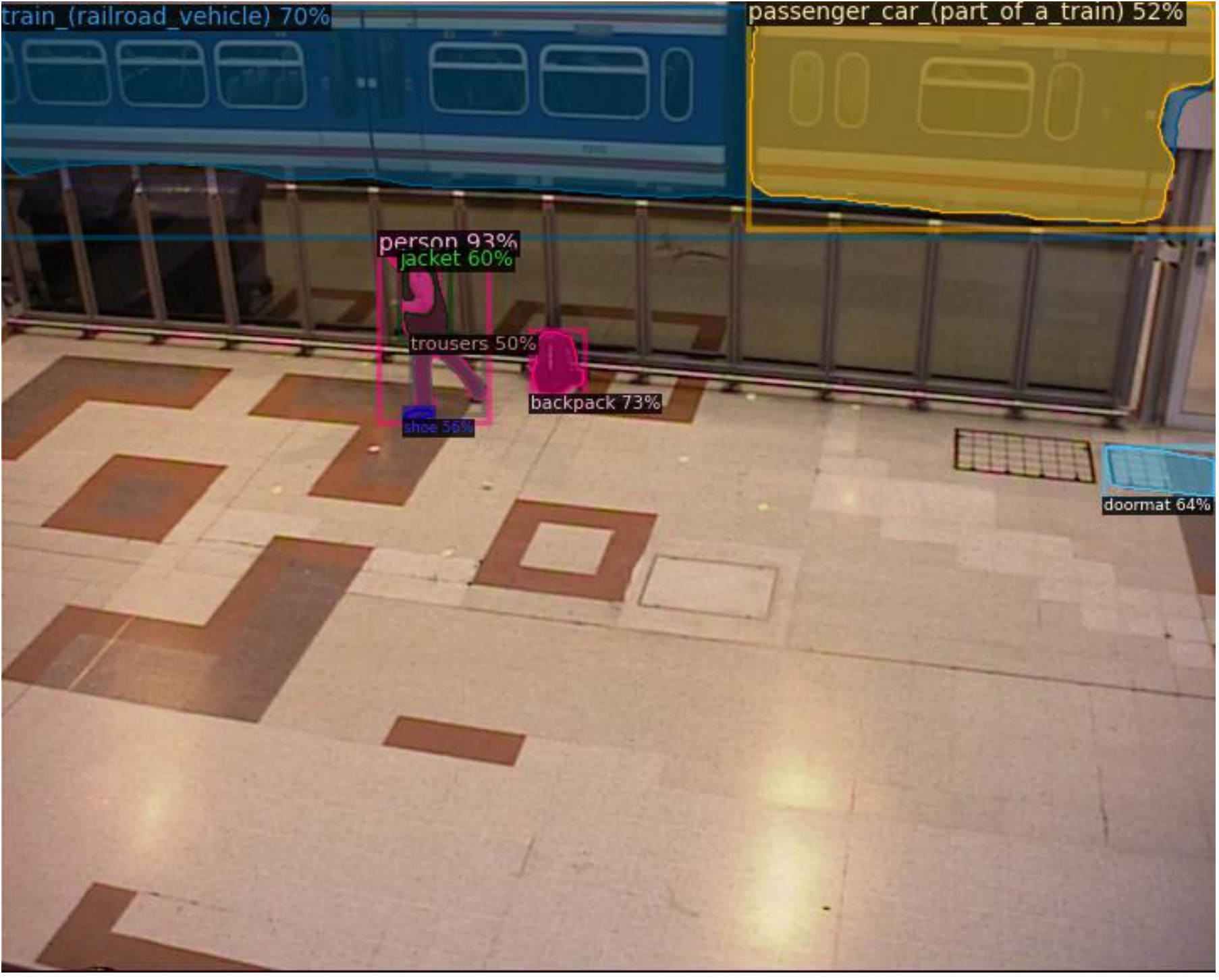}
        \caption{}
    \end{subfigure}
    \begin{subfigure}{0.32\linewidth}
        \includegraphics[width=1\linewidth]{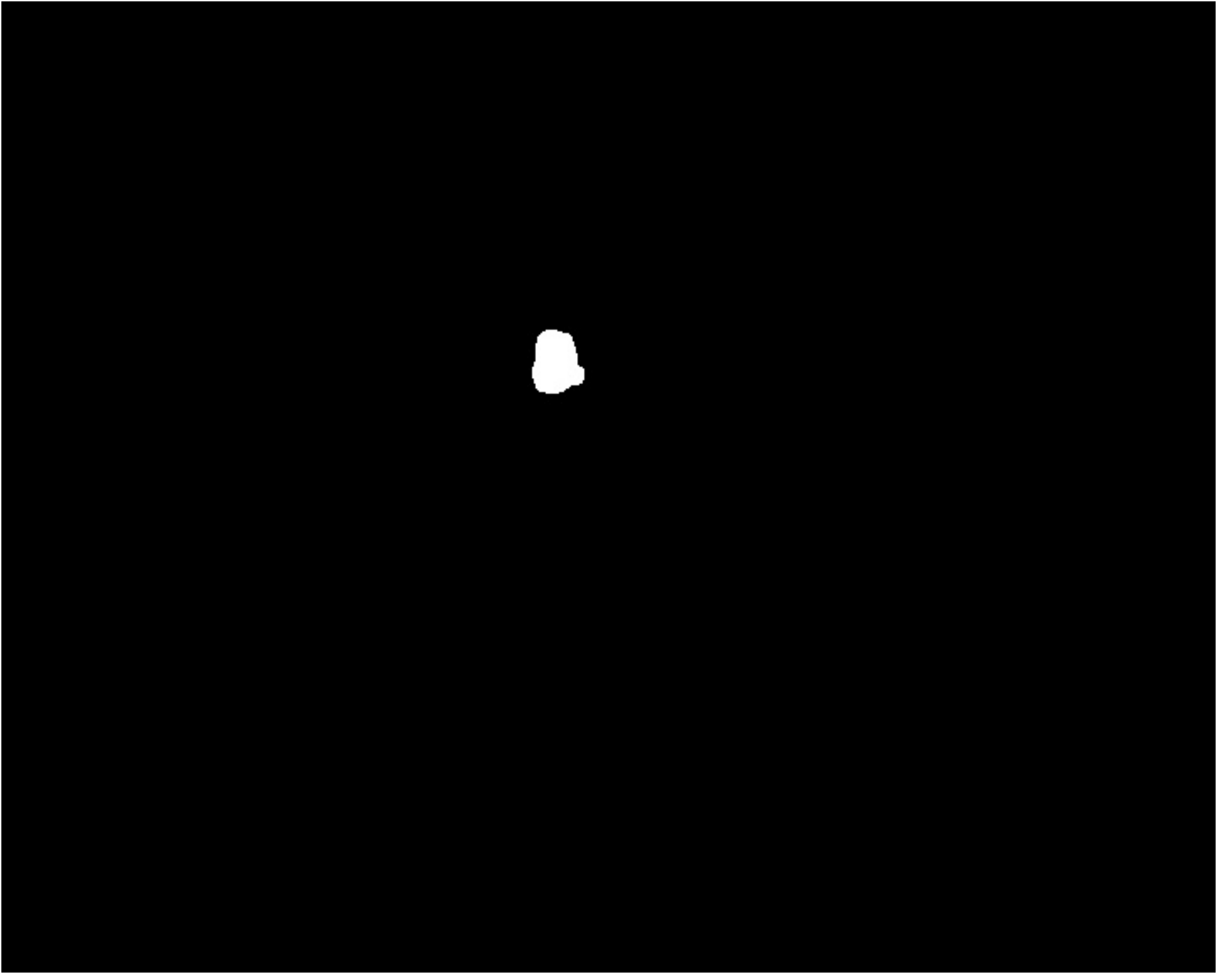}
        \caption{}
    \end{subfigure}
    \caption{The detection results of the PETS2006. (a) is the original frame. (b) is the all-instance detection results. (c) is the abandoned object detection results of our method.}
    \label{fig:pets}
\end{figure}
\vspace{-0.4cm}

\subsection{ABODA}

The $\textbf{AB}$andoned $\textbf{O}$bjects $\textbf{DA}$taset (ABODA) \cite{KevinLin2015AbandonedOD} contains 11 sequences that present a range of challenging scenarios for abandoned object detection, including crowded scenes, changes in illumination, night-time detection, and both indoor and outdoor environments. \Cref{fig:aboda} displays the results from the $video1.avi$ sequence in ABODA.

\vspace{-0.2cm}
\begin{figure}[ht]
    \centering
    \begin{subfigure}{0.32\linewidth}
        \includegraphics[width=1\linewidth]{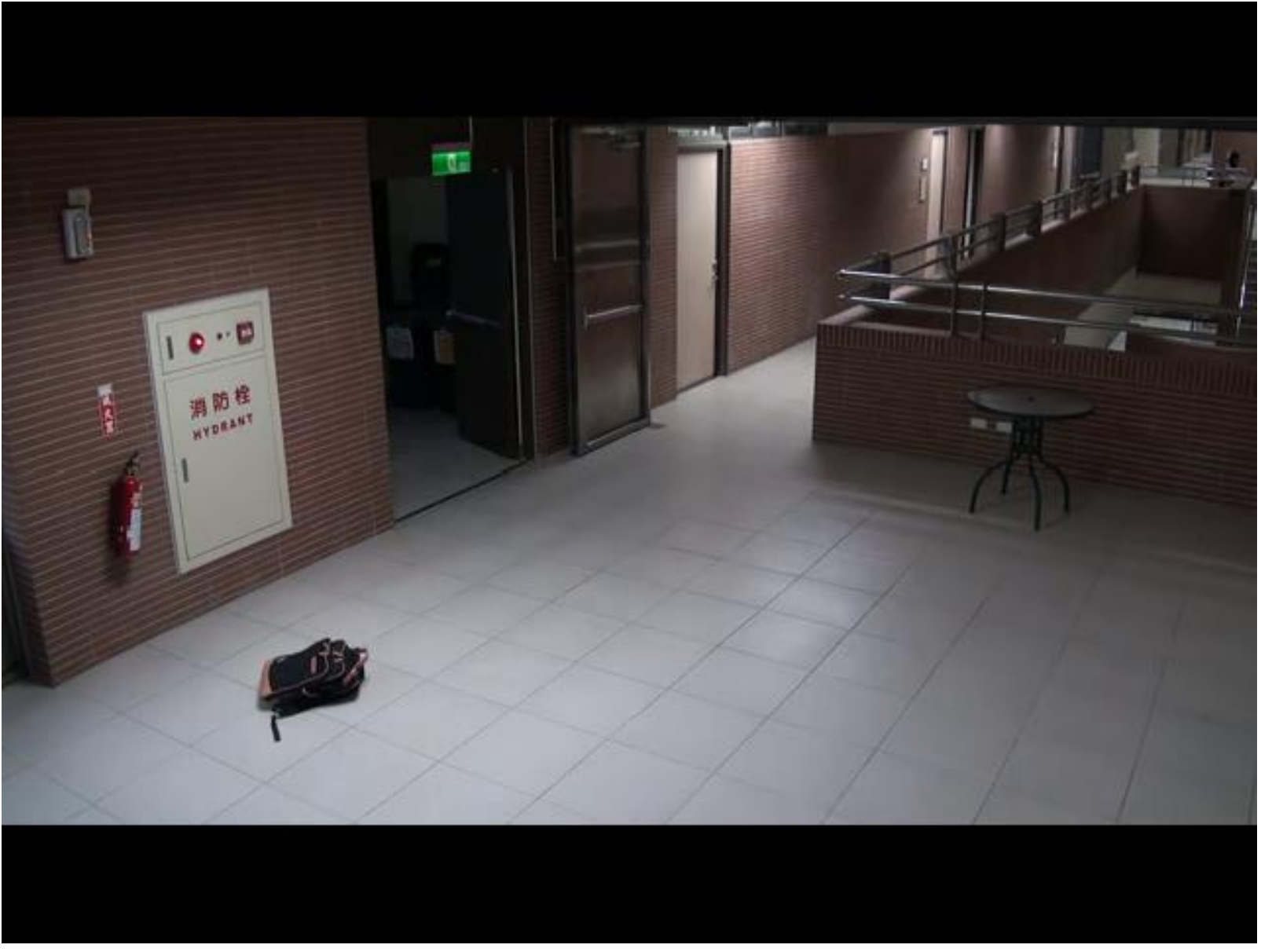}
        \caption{}
    \end{subfigure}
    \hfill
    \begin{subfigure}{0.32\linewidth}
        \hspace{-0.15cm}
        \includegraphics[width=1\linewidth]{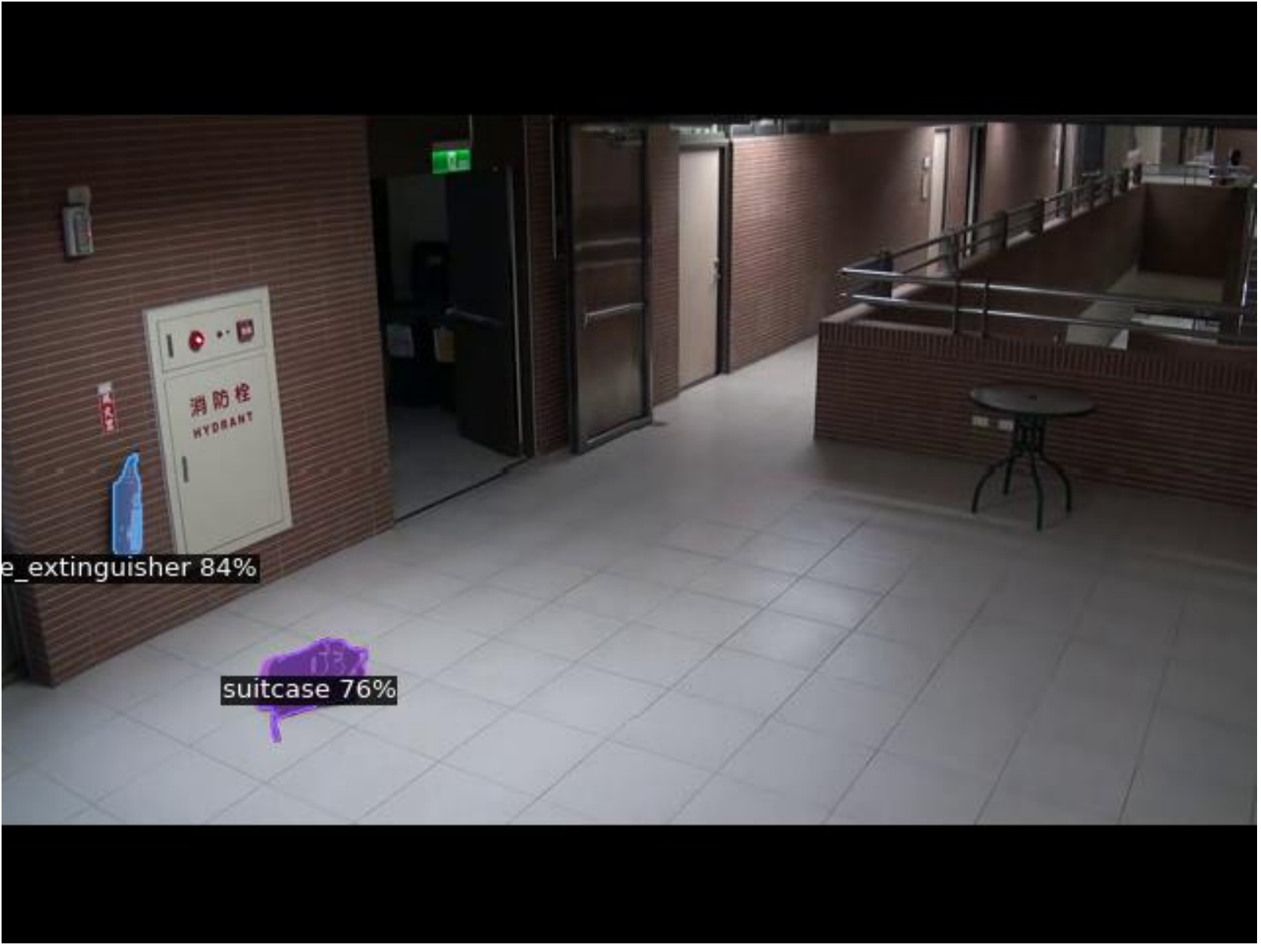}
        \caption{}
    \end{subfigure}
    \begin{subfigure}{0.32\linewidth}
        \includegraphics[width=1\linewidth]{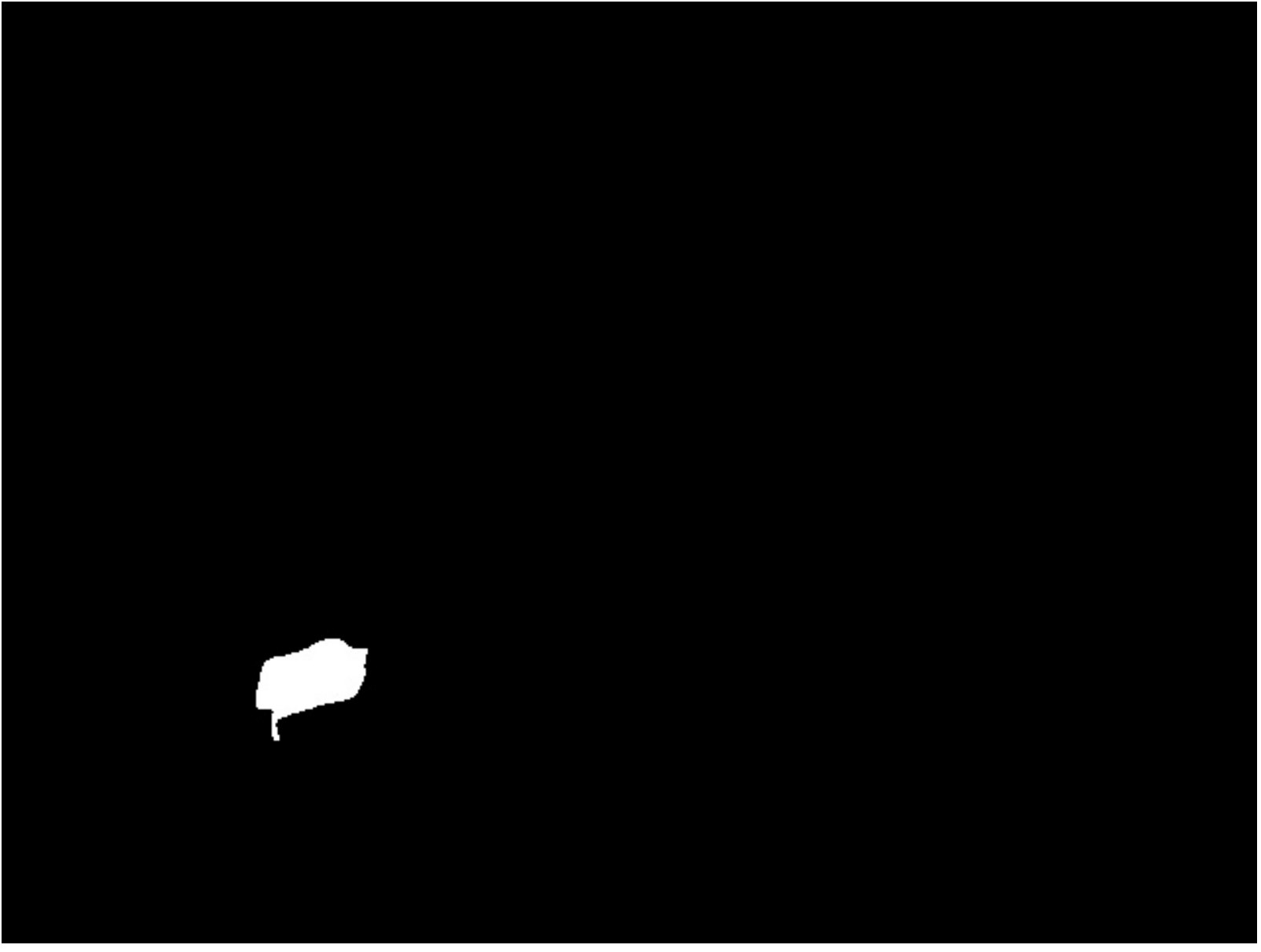}
        \caption{}
    \end{subfigure}
    \caption{The detection results of the ABODA. (a) is the original frame in the video. (b) is the all-instance detection results. (c) is the abandoned object detection results of our method.}
    \label{fig:aboda}
\end{figure}
\vspace{-0.4cm}

\subsection{TADA}

The $\textbf{T}$raffic $\textbf{A}$bandoned object detection $\textbf{DA}$taset (TADA) is a household traffic abandoned object detection dataset that comprises 20 sequences, 14 of which contain traffic abandoned objects. These objects typically consist of various types of traffic litter, such as plastic bags, which have diverse appearances and shapes and are usually carried by the wind. This presents a formidable challenge for abandoned object detection. \Cref{fig:tada} displays the results from the TADA dataset.

\vspace{-0.2cm}
\begin{figure}[ht]
    \centering
    \begin{subfigure}{0.32\linewidth}
        \includegraphics[width=1\linewidth]{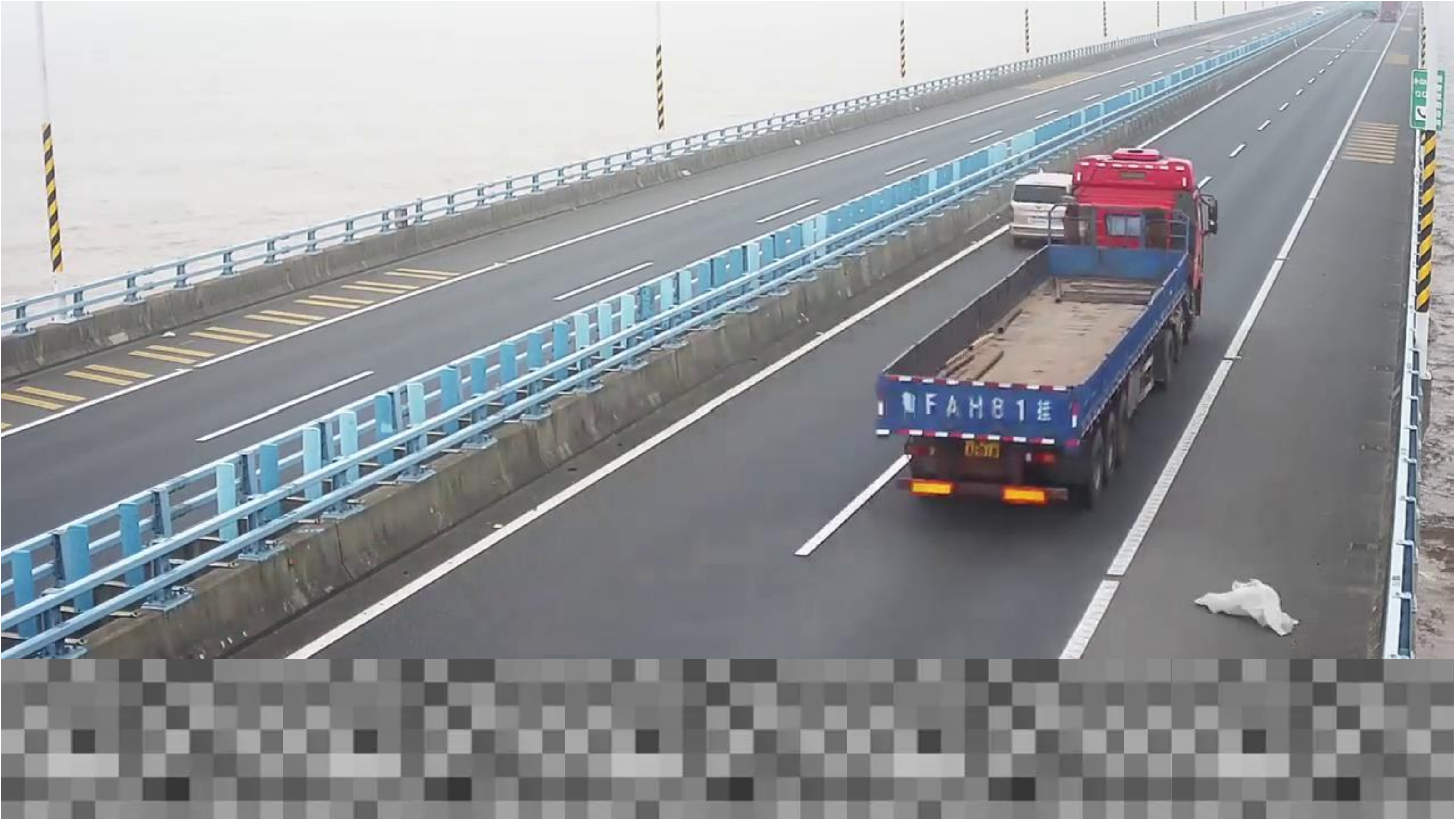}
        \caption{}
    \end{subfigure}
    \hfill
    \begin{subfigure}{0.32\linewidth}
        \hspace{-0.15cm}
        \includegraphics[width=1\linewidth]{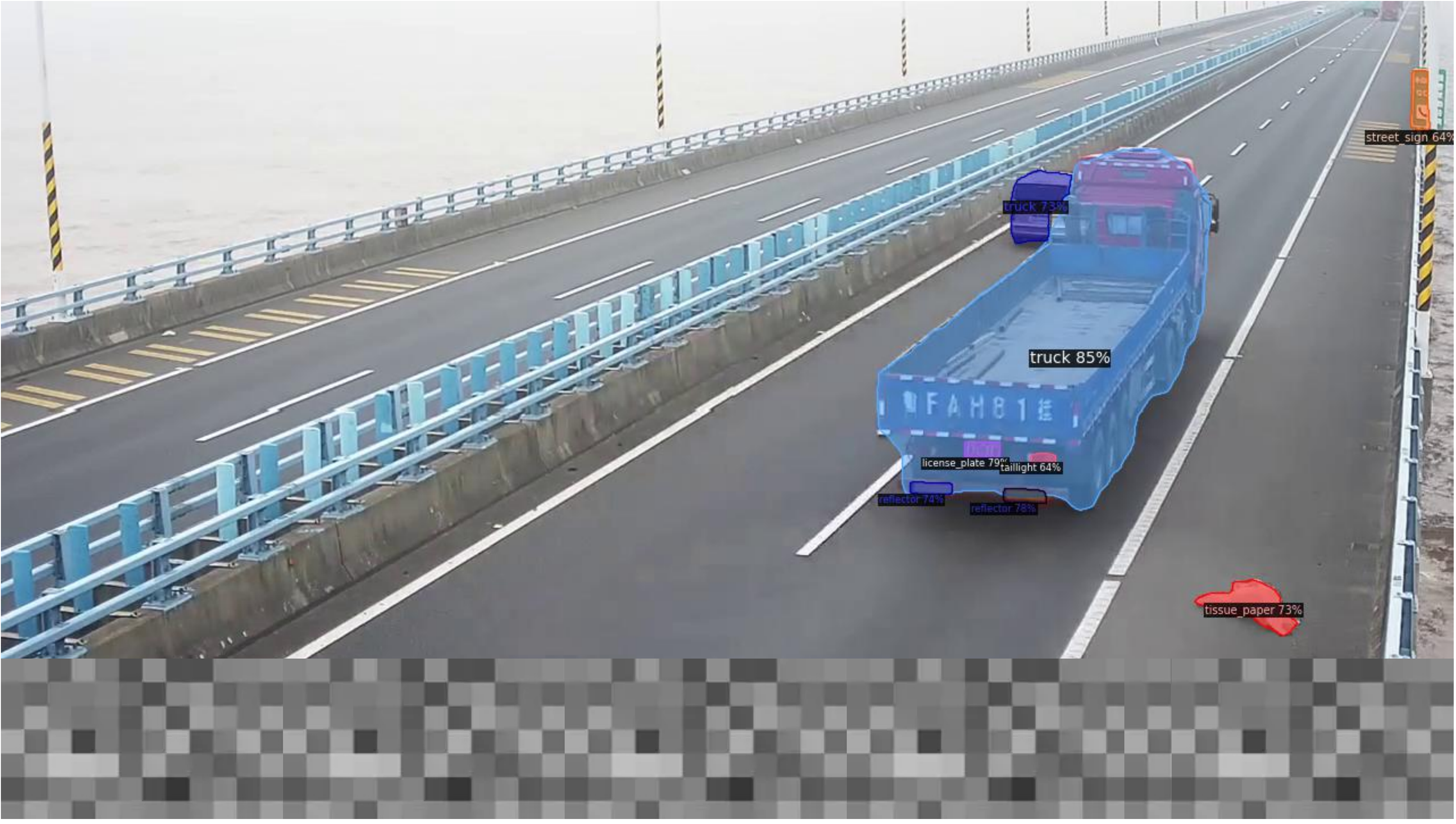}
        \caption{}
    \end{subfigure}
    \begin{subfigure}{0.32\linewidth}
        \includegraphics[width=1\linewidth]{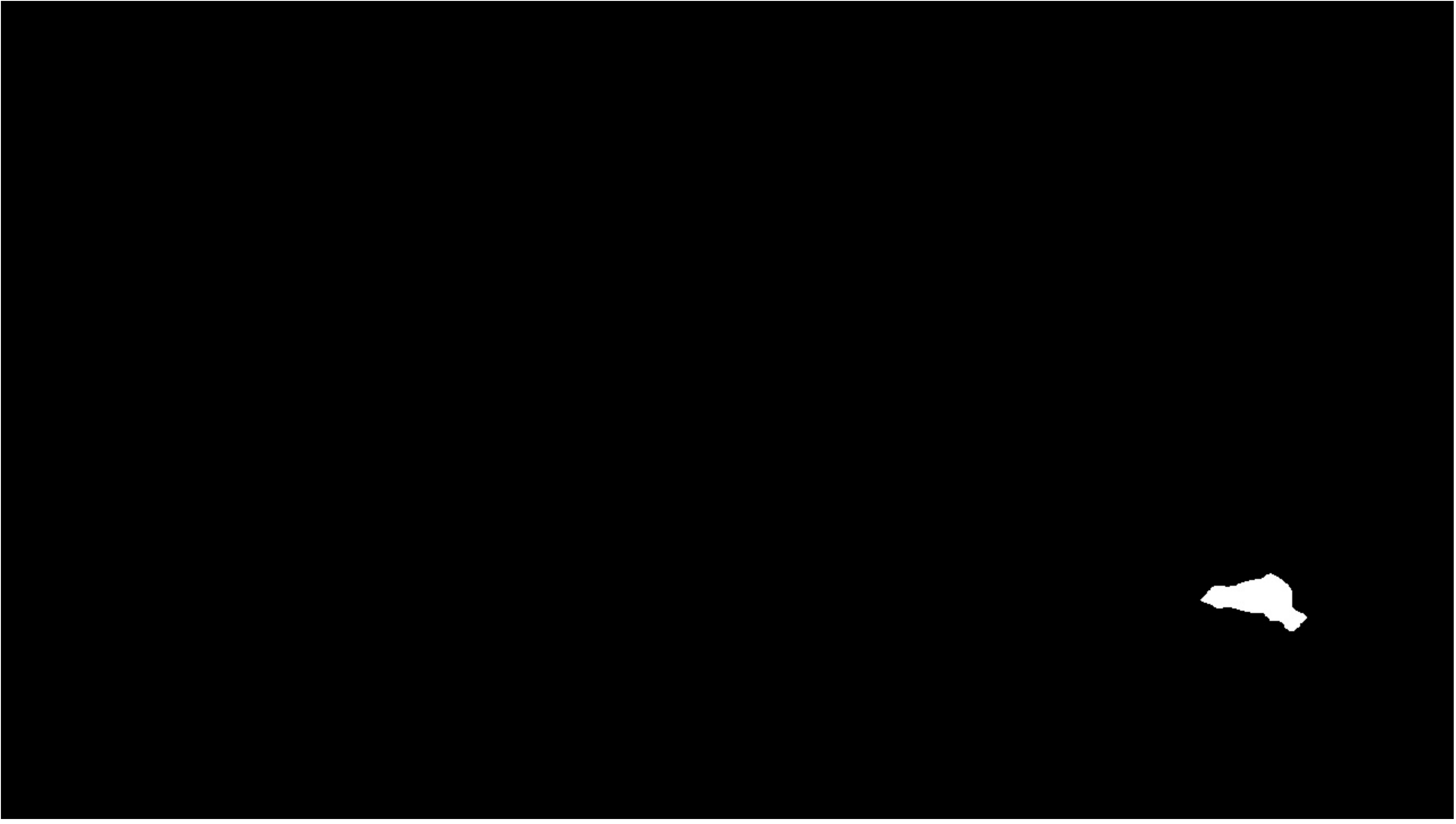}
        \caption{}
    \end{subfigure}
    \caption{The detection results of the TADA. (a) is the original frame in the video. (b) is the all-instance detection results. (c) is the abandoned object detection results of our method.}
    \label{fig:tada}
\end{figure}
\vspace{-0.4cm}

\end{document}